\documentclass[journal]{IEEEtran}
\usepackage{graphicx}
\usepackage{float}
\usepackage[caption=false]{subfig}
\usepackage{amsmath, amsfonts}
\usepackage{algorithm, algorithmic}
\usepackage{textcomp}
\usepackage{stfloats}
\usepackage{url}
\usepackage{bbm}
\usepackage{verbatim}
\usepackage{mathrsfs}
\usepackage{caption}
\usepackage{amssymb}
\usepackage{booktabs}
\usepackage{xcolor}
\usepackage{multirow}
\usepackage{hyperref}
\usepackage{listings}
\usepackage{enumerate}
\usepackage[normalem]{ulem}
\usepackage[percent]{overpic}
\usepackage[export]{adjustbox}
\usepackage{balance}
\newcommand{\czh}[1]{{#1}}

\newcommand{\eat}[1]{}

\newcommand{\name}{ECL}
\newcommand{\myparagraph}[1]{\vspace{2mm}\noindent\textbf{#1}}
\newcommand{\good}[1]{\textcolor[RGB]{55, 146, 55}{\scriptsize{\textbf{#1}}}}

\useunder{\uline}{\ul}{}
\def\BibTeX{{\rm B\kern-.05em{\sc i\kern-.025em b}\kern-.08em
    T\kern-.1667em\lower.7ex\hbox{E}\kern-.125emX}}

\begin{document}
\twocolumn
\title{Towards Effective Collaborative Learning in Long-Tailed Recognition}
\author{   
    Zhengzhuo Xu$^*$,
    Zenghao Chai$^*$,
    Chengyin Xu, \\
    Chun Yuan$^\dagger$, \textit{Senior Member, IEEE}, Haiqin Yang$^\dagger$, \textit{Senior Member, IEEE}
    \thanks{$^\dagger$Corresponding authors: C. Yuan is with the Tsinghua Shenzhen International Graduate School, Tsinghua University, Shenzhen 518055, China. (email: yuanc@sz.tsinghua.edu.cn).  H. Yang is with International Digital Economy Academy, Shenzhen 518045, China.  (email: hqyang@ieee.org).
    
    $^*$Equal contribution authors, listing order is random. Z. Xu, Z. Chai and C. Xu are with the Tsinghua Shenzhen International Graduate School, Tsinghua University, Shenzhen 518055, China. (e-mail: \{xzz20, xucy20\}@mails.tsinghua.edu.cn, zenghaochai@gmail.com).
    }
}

\maketitle
\begin{abstract}
    Real-world data usually suffers from severe class imbalance and long-tailed distributions, where minority classes are significantly underrepresented compared to the majority ones. Recent research prefers to utilize multi-expert architectures to mitigate the model uncertainty on the minority, where collaborative learning is employed to aggregate the knowledge of experts, i.e., online distillation. In this paper, we observe that the knowledge transfer between experts is imbalanced in terms of class distribution, which results in limited performance improvement of the minority classes. To address it, we propose a re-weighted distillation loss by comparing two classifiers' predictions, which are supervised by online distillation and label annotations, respectively. We also emphasize that feature-level distillation will significantly improve model performance and increase feature robustness. Finally, we propose an Effective Collaborative Learning ({\name}) framework that integrates a contrastive proxy task branch to further improve feature quality. Quantitative and qualitative experiments on four standard datasets demonstrate that {\name} achieves state-of-the-art performance and the detailed ablation studies manifest the effectiveness of each component in {\name}.
\end{abstract}
    

\section{Introduction}
\label{sec:intro}

\IEEEPARstart{R}{ecent} advancements in computer vision, e.g., visual recognition~\cite{ForestDet}, video analysis~\cite{VideoLT} and person re-ID~\cite{LTReID, IPDLT}, heavily rely on the large-scale, high-quality, and balanced datasets, such as ImageNet~\cite{Imagenet}, COCO~\cite{COCO} and Place~\cite{PlaceLT}, which require laborious collections and careful annotations.
Regrettably, collecting rare instances entails gathering more dominant samples because real-world data naturally exhibits imbalanced distributions w.r.t. its categories.
Hence, datasets typically follow a long-tailed distribution, with only a few labels having a majority of the samples, while most labels are associated with limited instances.
In Long Tail Recognition (LTR), the minority classes (\textbf{tail}) are always overwhelmed by the majority classes (\textbf{head}), resulting in low performance for the tail. As a result, the models trained on the long-tailed dataset show great uncertainty, where the outputs for few-shot classes vary remarkably, despite the same training settings.

\begin{figure}[t!]
	\centering
	\subfloat[Collaborative Learning]{
        \includegraphics[width=0.45\linewidth]
        {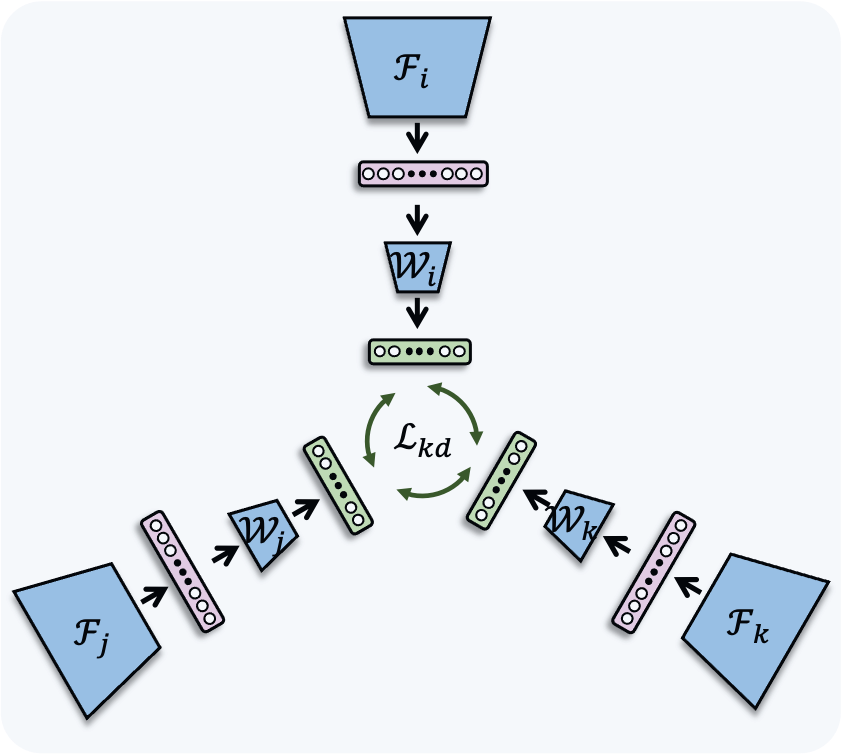}
        \label{fig:teaser_ncl}
    }
    \subfloat[ECL (Ours)]{
        \includegraphics[width=0.45\linewidth]
        {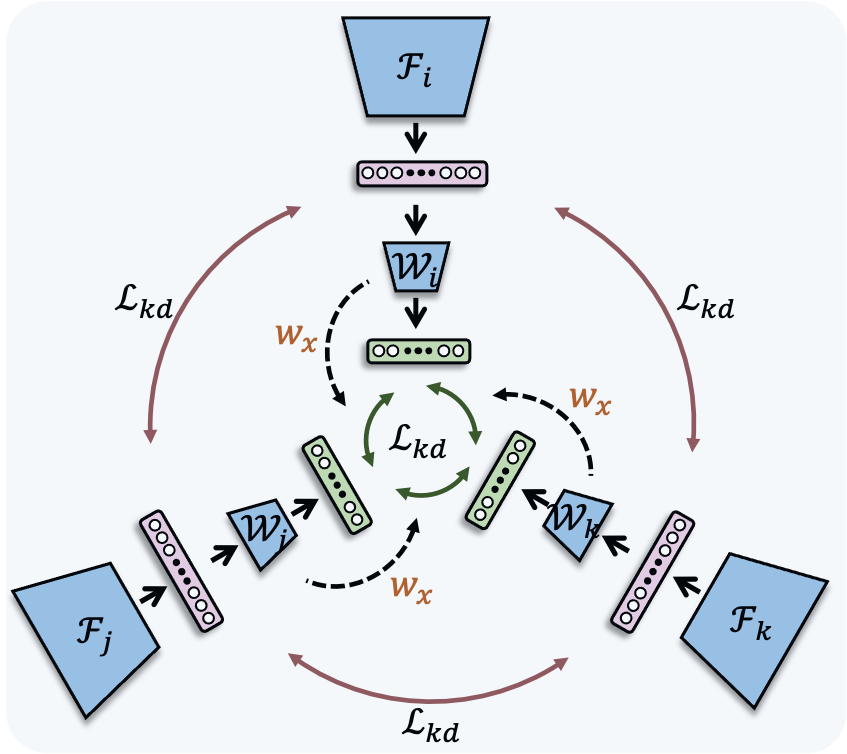}
        \label{fig:teaser_ecl}
    }
	\caption{The illustration of collaborative learning in the multi-expert framework. $\mathcal{F}$: feature encoder. $\mathcal{W}$: classification head. Different from previous work, we re-balance the distillation and conduct online distillation on both feature and logit levels.}
	\label{fig:teaser}
\end{figure}

Most existing work addresses the LTR issue by improving the feature representations of tail classes or re-balancing the contribution of different classes. 
However, some intuitive approaches like over-sampling the tail~\cite{BBN} or under-sampling the head~\cite{Bagoftricks} result in severe robustness problems especially in tail classes. Although some well-designed approaches enrich tail samples in more elegant ways, such as through feature combinations~\cite{Dynamicmix, PriorLT, FSA} or pseudo sample generation~\cite{M2m, Bagoftricks, CMO}, the problem of model preference towards head classes remains unresolved.
To calibrate the label distribution gap between the train and test dataset, the Balanced Cross-entropy (BC) loss is proposed based on \textit{Bayesian Theory}, which compensates the model bias by label frequency on standard \textit{softmax} Cross-Entropy (CE) loss~\cite{BS, LA, LADE, PriorLT}.
Based on the effective BC loss, Multi-Expert (ME)~\cite{LFME, RIDE, NCL} framework is proposed to further address model uncertainty on the tail classes. 
For example, NCL~\cite{NCL} trains several expert networks in parallel and aggregates each expert's knowledge in a nested collaborative manner, i.e., online Knowledge Distillation (KD) on the logit level (see Fig.~\ref{fig:teaser_ncl}), where we refer to each network as an expert.

However, our experimental observations indicate that the transfer knowledge (distillation logit value) is not balanced w.r.t. class in vanilla collaborative learning (see Sec.~\ref{sec:motivation}).
The tail samples are always under-represented during the distillation process, which damages the balanced knowledge transfer.
Such imbalance leads the online distillation to boost the head performance while suppressing the transfer of tail knowledge. Consequently, the tail remains unimproved compared to the single expert baseline.
Recent research~\cite{RTG} suggests that the KD-trained classifier is more confident for the over-represented samples than the label-trained one because the distillation tends to learn more generalized \textit{context} knowledge compared to label supervision, which mainly provides content-invariant knowledge.
Inspired by it, we propose a novel re-weighted distillation loss by comparing the predictions of two different classifiers.
Moreover, we propose to perform additional collaborative distillation at the feature level, which significantly boosts model performance and feature robustness.
We further incorporate a contrastive proxy task with a parallel branch to improve feature representations.
As a result, we propose a novel Effective Collaborative Learning ({\name}) framework to improve vanilla NCL, which distinguishes from previous ME frameworks in two aspects:

\textbf{\textit{Single expert training.}}
We propose the Balanced Knowledge Transfer (BKT) module to conduct balanced knowledge distillation. 
Following the feature encoder, we add an extra reference classifier parallel to the original classifier. The reference classifier is only supervised by the BC loss and is not involved in the expert collaboration, allowing it to only focus on the content-invariant knowledge. We compare the predictions of two classifiers to estimate whether the input samples are over-confident or not and re-weight the KD loss to assign the under-represented samples with larger weights (Fig.~\ref{fig:teaser_ecl}).
For each expert, we introduce a siamese branch to conduct Contrastive Proxy Task (CPT) and update parameters in a momentum-based moving average scheme~\cite{MoCo}. The CPT is designed to increase the feature similarity of an image's two views to facilitate model discriminative ability.
Note that we will discard the additional reference classifier and siamese branch during the inference phase to keep the consistent architecture with previous ME approaches.

\textbf{\textit{Expert knowledge aggregation.}}
In the proposed {\name}, each expert is collaboratively learned with others. Note that the knowledge is transferred not only on the logit level but also on the feature level (see Fig.~\ref{fig:teaser_ecl}), which facilitates stable representation learning. Our Feature Level Distillation (FLD) is a simple yet effective improvement that encourages all experts to extract well-represented features. We also present in-depth analysis to investigate how FLD influences the model performance qualitatively and quantitatively (see Sec.~\ref{sec:exp}).

With the above observations, insights, and techniques, we build our final {\name} (Fig.~\ref{fig:teaser_ecl}\&\ref{fig:single_pipeline}), which contains three key components, namely the balanced knowledge transfer module, feature level distillation, and contrastive proxy task. Extensive experiments in four benchmarks justify the superiority of {\name}. In summary, our contributions are as follows:
\begin{enumerate}
\item[1)] We pinpoint the imbalance of transfer knowledge in previous collaborative learning methods and propose a balanced knowledge distillation loss to tackle it.
\item[2)] We propose to conduct knowledge distillation on both feature and logit levels, which significantly enhances model performance and robustness. 
\item[3)] We propose the {\name} framework to collaboratively train multiple experts to overcome the head preference and tail uncertainty in long-tailed recognition.
\item[4)] We present extensive experiments and demonstrate {\name} achieves state-of-the-art performance on CIFAR10/100-LT, ImageNet-LT, and iNaturalist 2018 datasets.
\end{enumerate}

This paper is organized as follows: Sec.~\ref{sec:relate} provides a brief overview of related work. In Sec.~\ref{sec:prelim}, we introduce the relevant concepts and baselines. We discuss our motivation based on experimental observations in Sec.~\ref{sec:motivation} and provide a detailed design in Sec.~\ref{sec:method}. Sec.~\ref{sec:exp} demonstrates the effectiveness of {\name} through extensive experiments and ablation studies. Finally, Sec.~\ref{sec:conclusion} concludes our work.

1. We pinpoint the imbalance of transfer knowledge in previous collaborative learning methods and propose a balanced knowledge distillation loss to tackle it.
2. We propose to conduct knowledge distillation on both feature and logit levels, which significantly enhances model performance and robustness. 
3. We propose the ECL framework to collaboratively train multiple experts to overcome the head preference and tail uncertainty in long-tailed recognition.
4. We present extensive experiments and demonstrate ECL achieves state-of-the-art performance on CIFAR10/100-LT, ImageNet-LT, and iNaturalist 2018 datasets.

\section{Related Work}
\label{sec:relate}

\noindent\textbf{Feature-wise Rebalance Learning.} 
To avoid damaging model generalization severely from simply over/under-sampling the tail/head classes~\cite{oversample1, NCM, Bagoftricks}, recent advances resort combination of the head to enrich the feature of tail samples~\cite{M2m, Bagoftricks, RSG} or increase the tail frequency implicitly~\cite{Dynamicmix, GeneticGAN, PriorLT, CMO}. The two-stage methods~\cite{LDAM, NCM, MiSLAS} decouple feature learning from downstream tasks (e.g., classification) to reduce the bias on the classifier. Several methods~\cite{SimCLR, MoCo, SSLMM} also leverage self-supervised learning to eliminate the influence of imbalanced distribution. SSP~\cite{SSP} and HybirdSC~\cite{HybirdSC} demonstrated that self-supervised or semi-supervised training can boost performance through larger train epochs and GPU memory. Recent state-of-the-art~\cite{PaCo, HybirdSC, TSC, BCL} introduces fixed or learnable proxy to overcome performance degradation due to the absence of label supervision.

\noindent\textbf{Reweight-wise Learning.} 
To mitigate the inherent statistical bias in LTR, researchers have designed meticulous loss to learn larger \textit{margins} among different classes~\cite{LDAM, LA, LADE, BS, PriorLT, DAP, ISFDA, GCL} or assign various \textit{weights} for different classes based on the label frequencies~\cite{CB, LTDA, EqLoss, MiSLAS, LTR-WD}. In particular, the simple yet effective BC loss~\cite{LA, BS, LADE, PriorLT} has been widely adopted by state-of-the-art~\cite{NCL, PaCo, BCL, TADE}. Unfortunately, BC loss is not always compatible with the above feature-wise methods for the inconsistency of the statistical label frequency.

\noindent\textbf{Multi-expert Learning.}
To tackle the tail uncertainty~\cite{CBD, RIDE}, the multi-expert framework is increasingly valued, which typically contains two components, i.e., \textit{single expert training} and \textit{experts knowledge aggregation}~\cite{BBN, ACE, LFME, RIDE, TADE}. BBN~\cite{BBN} trains two experts with instance sampling and inverse sampling, respectively, and aggregates their knowledge in a cumulative weighting manner. LFME~\cite{LFME} trains multiple experts with different instance groups and weights the logits from different experts as the final output. RIDE~\cite{RIDE} enlarges the KL divergence to train experts and cascades all the experts via decision gates for inference. TADE~\cite{TADE} trains experts by BC loss with different assumed statistical prior and weights each expert's output, which is obtained via post-hoc contrastive training. Another feasible expert aggregation manner is knowledge distillation~\cite{KD}. DiVE~\cite{DiVE} shows the effectiveness of distillation in the LTR. SSD~\cite{SSD} trains expert backbone by self-distillation learning and classifier through balanced sampling. CBD~\cite{CBD} trains different teachers by various data augmentations and random seeds. Then, it trains students with balance sampling and knowledge from the above teachers. NCL~\cite{NCL} trains experts in a nested manner and adopts online inter-distillation with each other to reduce the tail uncertainty. However, these methods mainly conduct logit-level distillation while ignoring the imbalance of transfer knowledge.
\section{Preliminaries}
\label{sec:prelim}

\subsection{Task Definition.}
{Given an $N$-sample dataset $\mathcal{D}=\{(\mathbf{x}_i , \mathbf{y}_i)\}_{i=1}^N$ from $C$ classes, where $\mathbf{x}_i \in \mathcal{X}$ denotes the $i$-th instance with its label, $\mathbf{y}_i \in \mathcal{Y}:=\{\mathbf{y}_1,\ldots,\mathbf{y}_C\}$.} We assume the dataset $\mathcal{D}$ is long-tailed distributed and denote each category as $\mathcal{C}_i$ and its instance number as $n_i=|\mathcal{C}_i|$. Furthermore, we consider a base classification model $\mathcal{M}:=\{\mathcal{F}_{\theta}, \mathcal{W}_{\phi}\}$. It contains a learnable \textit{feature encoder} $\mathcal{F}_{\theta}$ and a \textit{classifier} $\mathcal{W}_{\phi}$, parameterized by $\theta$, $\phi$, respectively. Given an input image $\mathbf{x}$, the encoder extracts the feature representation $\mathbf{v}:=\mathcal{F}_{\theta}(\mathbf{x})\in \mathbb{R}^d$. Then, the classifier (typically fully connected layers) outputs the logits $\mathbf{z}:=\mathcal{W}_{\phi}(\mathbf{v}) \in \mathbb{R}^C$. We assume $K$ experts in the collaborative learning framework with the same architecture $\mathcal{M}$ and denote the $k$-th expert as ${E}_k:=\{\mathcal{F}_{\theta_k}, \mathcal{W}_{\phi_k}\}$.

\subsection{Balanced Cross-entropy Loss.}
\textbf{B}alanced \textbf{C}ross-entropy (BC) loss is effective and widely adopted in LTR tasks~\cite{LA, BS, LADE, PriorLT, NCL, BCL}. It compensates the statistical bias via logits adjustment on standard \textbf{C}ross-\textbf{E}ntropy (CE) loss. Consider the expert $E_k$ is supervised by CE loss with standard \textit{softmax}:

\begin{equation}
\label{Eq.CE}
    \begin{aligned}
        \mathcal{L}_{\text{CE}} = - \log\left(p(\mathbf{y}_i|\mathbf{x};\theta_k,\phi_k)\right) 
        = \log\left[1+\sum_{\mathbf{y}_j\neq \mathbf{y}_i} e^{\mathbf{z}_{\mathbf{y}_j} - \mathbf{z}_{\mathbf{y}_i}}\right].
    \end{aligned}
\end{equation}

Here, we denote the label distribution prior of train/test data as $p_s(\mathbf{y})$/$p_t(\mathbf{y})$ respectively. Based on the Bayesian theory, \textit{the posterior is proportional to prior times likelihood}, where the likelihood $p_s(\mathbf{x}|\mathbf{y})$ maximization is equal to the model parameters (i.e., $\theta, \phi$) learning. Typically, the posterior $p_t(\mathbf{y}|\mathbf{x})$ is equivalent to likelihood $p_s(\mathbf{x}|\mathbf{y})$ between train and test set when $p_s(\mathbf{y})\equiv p_t(\mathbf{y})$. However, if we take the statistical distribution of label $\mathbf{y}$ as its prior $p(\mathbf{y})$, we can derive the following bias from the mismatch of $p_s(\mathbf{y})$ and $p_t(\mathbf{y})$:
\begin{equation}
\label{Eq.bias}
    \begin{aligned}
        &p_t(\mathbf{y}|\mathbf{x}) = \frac{p_s(\mathbf{x}|\mathbf{y}) \cdot p_s(\mathbf{x})}{p_s(\mathbf{y})}\cdot \frac{p_t(\mathbf{y})}{p_t(\mathbf{x})} \propto \frac{p_s(\mathbf{x}|\mathbf{y})\cdot p_t(\mathbf{y})}{p_s(\mathbf{y})} \\
        &= \frac{\frac{p_t(\mathbf{y})}{p_s(\mathbf{y})} \cdot e^{\mathbf{z}_{\mathbf{y}_i}}}{\sum_{j} \frac{p_t(\mathbf{y}_j)}{p_s(\mathbf{y}_j)} \cdot e^{\mathbf{z}_{\mathbf{y}_j}}} = \frac{e^{\mathbf{z}_{\mathbf{y}_i} {\color{red}-} (\log(p_s(\mathbf{y}_i)) - \log(p_t(\mathbf{y}_i)) ) }}{\sum_j e^{\mathbf{z}_{\mathbf{y}_j} {\color{red}-} ( \log(p_s(\mathbf{y}_j)) - \log(p_t(\mathbf{y}_j)) ) }},
    \end{aligned}
\end{equation}
where $p_s(\mathbf{x})$ and $p_t(\mathbf{x})$ are regular terms (i.e., normal distribution). Here, we get statistical bias of class $\mathbf{y}_i$ as $\log p_s(\mathbf{y}_i) - \log p_t(\mathbf{y}_i)$. Combining Eq.~\ref{Eq.CE} and Eq.~\ref{Eq.bias}, we compensate for it in CE loss with a hyper-parameter $\tau$ as follows:
\begin{equation}
\label{Eq.bc_loss}
    \mathcal{L}_{\text{BC}} = -\log\left[\frac{e^{\mathbf{z}_{\mathbf{y}_i} {\color{red}+} \tau \cdot \left( \log(p_s(\mathbf{y}_i)) - \log(p_t(\mathbf{y}_i)) \right) }}{\sum_j e^{\mathbf{z}_{\mathbf{y}_j} {\color{red}+} \tau \cdot \left(\log(p_s(\mathbf{y}_j)) - \log(p_t(\mathbf{y}_j)) \right)}}\right].
\end{equation}

\subsection{Nested Collaborative Learning}
To reduce the great uncertainty in long-tailed learning, Li \textit{et al}.~\cite{NCL} propose Nested Collaborative Learning (NCL) to learn multiple experts parallelly and aggregate the expert knowledge via nested online distillation on the logit-level (see Fig.~\ref{fig:teaser_ncl}). The NCL performs online inter-distillation on both partial and full views, while incorporating an instance discrimination task as well. All experts adopt the same BC loss and hyper-parameter settings. It contributes to complementary expert learning and achieves state-of-the-art performance whether by using a single expert or an ensemble. Our {\name} is motivated by the experimental observations on it, which will be elaborated in the following section.
\section{Motivation}
\label{sec:motivation}

\begin{figure}[t!]
	\centering
	\subfloat[Collaborative Learning]{
        \includegraphics[width=0.45\linewidth]
        {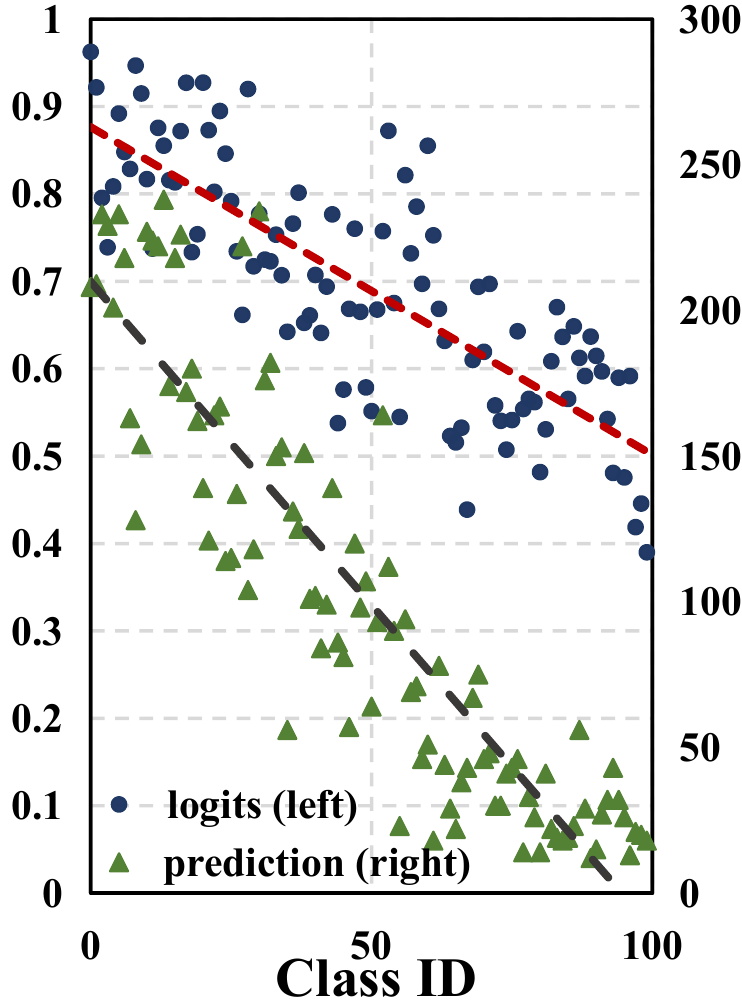}
        \label{fig:motivation_ncl}
    }
    \subfloat[ECL (Ours)]{
        \includegraphics[width=0.45\linewidth]
        {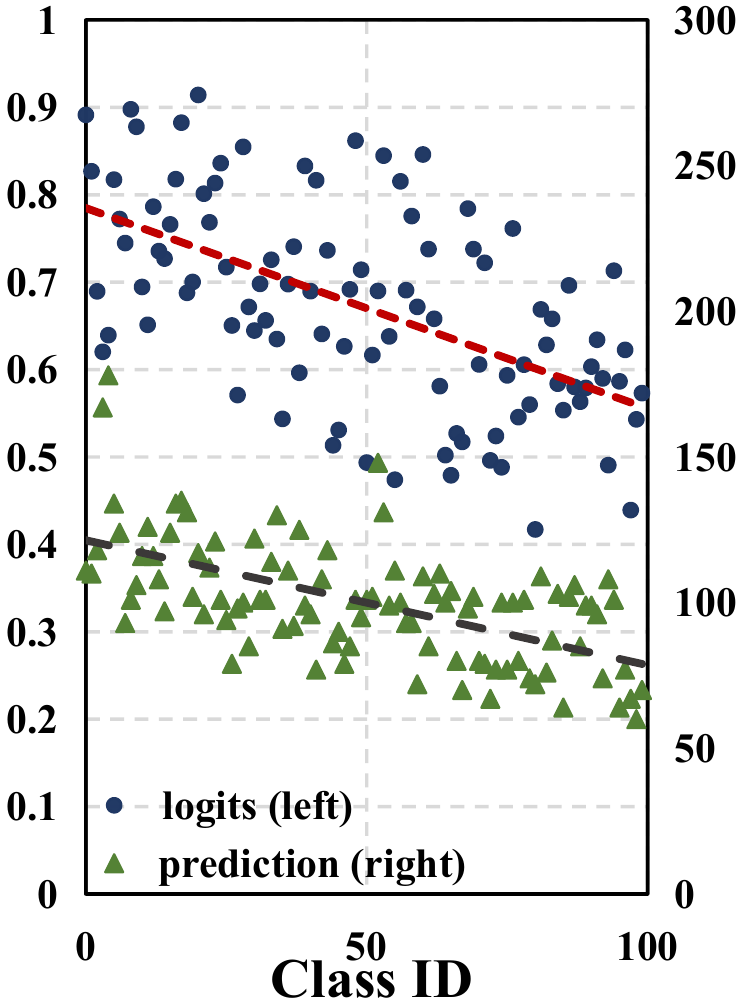}
        \label{fig:motivation_ecl}
    }
	\caption{Average distillation \textit{logits value} and \textit{prediction distribution} w.r.t. classes. We conduct an evaluation of vanilla NCL and proposed {\name} on CIFAR100-LT ($\gamma=100$). The class index is ranked according to the training instance number.}
	\label{fig:motivation}
\end{figure}

Our motivation stems from the following inspiring observations \textbf{1}: \textit{The transfer knowledge during online distillation is imbalanced w.r.t. classes in vanilla nested collaborative learning.} \textbf{2}: \textit{The optimal hyper-parameter of Eq.~\ref{Eq.bc_loss} for each expert is not consistent, which hinders performance improvement.}

For multi-expert collaborative learning approaches, the expert knowledge aggregation typically conduct at the logit level. For observation \textbf{1}, we demonstrate that previous distillation at logit level (Fig.~\ref{fig:teaser}) is ineffective. In Fig.~\ref{fig:motivation_ncl}, we visualize the logit-level transfer knowledge (i.e., distilled logits value) and model prediction numbers w.r.t. class of vanilla collaborative learning and ours. The tail classes present lower knowledge weight and thus result in fewer predictions. Similar observations also occur when training with balanced datasets, where the imbalance correlates to both \textit{label} and \textit{context}~\cite{RTG}. While NCL eliminates the model preference by compensating the label frequency with BC loss, the content imbalance remains unsolved. In this paper, we propose a balanced distillation loss to manage the label and context distribution simultaneously. Fig.~\ref{fig:motivation_ecl} shows that our proposal ameliorates the transfer knowledge imbalance and model prediction preference remarkably.

For observation \textbf{2}, we conduct an in-depth analysis by implementing the BC loss of NCL in a post-hoc manner~\cite{LA, LADE, PriorLT} as a \textit{softmax} variation (Eq.~\ref{eq:post_bias}). On the CIFAR100-LT ($\gamma=100$), the best performance achieves at $\tau \approx 1$ ($1.2$, $1.3$ or others) instead of theoretical $\tau=1$. It suggests that the statistical bias learned by each expert is inaccurate and inconsistent, which can be amplified and distorted further if we conduct knowledge aggregation on the logit level. As a comparison, label distribution seldom affects feature-level distillation because it only works on the logit-level~\cite{NCM}. Hence, we conduct feature level distillation to overcome this inconsistency. We resort optimization-related diagnostic tools (e.g., average feature distance among experts and normalized loss landscapes~\cite{Landscape}) to further explore the feature distillation mechanism, and present the visualized results in Sec.~\ref{sec:exp}.
\begin{equation}
    p(\mathbf{y}_i|\mathbf{x};\theta,\phi) = \frac{e^{\mathbf{z}_{\mathbf{y}_i} {\color{red}-}\tau \cdot (\log(p_s(\mathbf{y}_i)) - \log(p_t(\mathbf{y}_i)) ) }}{\sum_j e^{\mathbf{z}_{\mathbf{y}_j} {\color{red}-} \tau \cdot ( \log(p_s(\mathbf{y}_j)) - \log(p_t(\mathbf{y}_j)) ) }}.
    \label{eq:post_bias}
\end{equation}

\section{Methodology}
\label{sec:method}

\begin{figure*}[t!]
	\centering
	\flushleft
	\begin{overpic}[trim=0cm 0cm 0cm 0cm,clip,width=1\linewidth,grid=False]{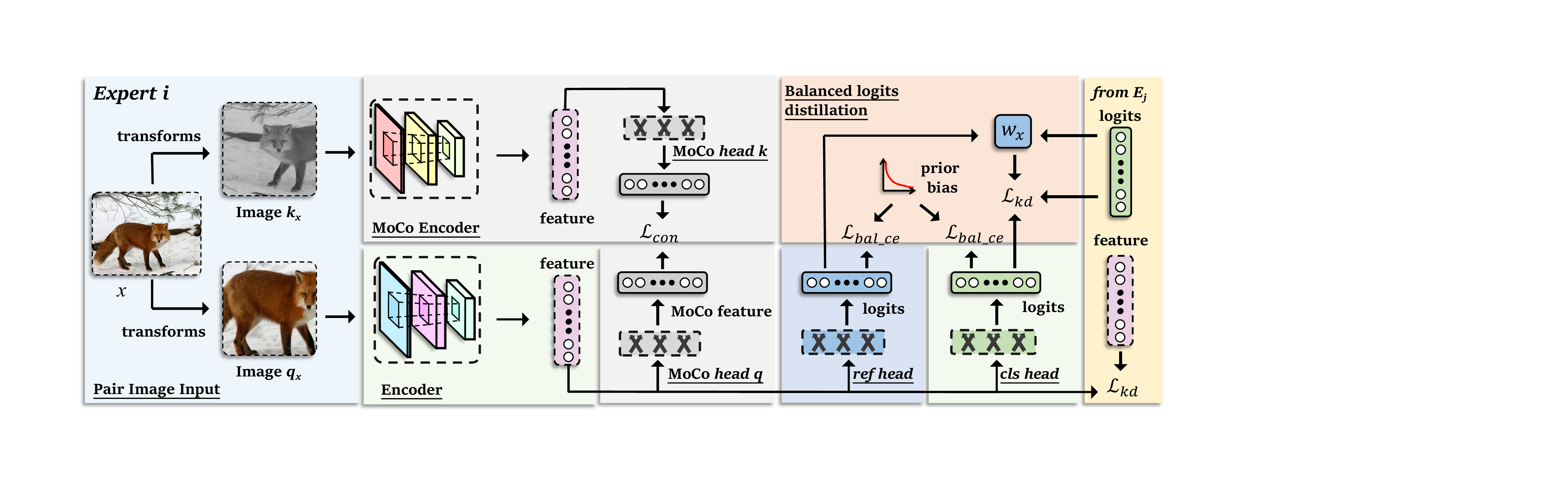}
    \end{overpic}
	\caption{The single expert architecture and corresponding training pipeline. Online distillation is performed at both the feature-level and logit-level. Auxiliary modules like \textit{ref} head, MoCo encoder \& head will be discarded during the inference phase.}
	\label{fig:single_pipeline}
\end{figure*}

In this section, we first propose a novel transfer module to rebalance the knowledge of each expert. Then we design a novel knowledge aggregation pipeline to manage the feature-wise collaboration and eliminate the statistical bias among multi-experts. We also leverage the unsupervised instance discrimination proxy to further boost the feature representation. Finally, we describe how to adjust the model for inference.

\subsection{Balanced Knowledge Transfer}
As discussed in Sec.~\ref{sec:motivation}, the imbalance is two-fold. While BC loss manages to eliminate the label imbalance by leveraging the distribution gap, it is challenging to eliminate the implicit context prior, which is not identically distributed with the labels.
Hence, we try to alleviate this issue from another perspective. In Fig.~\ref{fig:single_pipeline}, we propose the balanced knowledge transfer (BKT) module to tackle the imbalance issue when transferring the logit-level knowledge (Fig.~\ref{fig:motivation}). 

The proposed BKT is based on the experimental observations that \textit{the classifier supervised by knowledge distillation is more confident for the over-represented samples than the ones supervised by labels}~\cite{KD, RTG}. We attempt to identify underrepresented samples by comparing the predictions gap between two classifiers. Specifically, for the balanced distillation, we propose an extra reference head $\mathcal{W}_{\phi}^{r}$, which is only supervised by the BC loss and parallel to the vanilla classification head $\mathcal{W}_{\phi}^{c}$. With more soft supervision given by the KD loss, \textit{cls} head $\mathcal{W}_{\phi}^{c}$ learns more context equivariance knowledge compared to the \textit{ref} head $\mathcal{W}_{\phi}^{r}$, which only focuses on the context invariance knowledge (i.e., class labels). Therefore, BKT automatically identifies whether a sample is under-represented by comparing the decisions from the two heads:
\begin{equation}
    \hat{\mathbf{y}}^{r}_i=\frac{\exp \left(\mathbf{z}_{\mathbf{y}_i}^{r}/\sigma^{r}\right)}{\sum_{\mathbf{y}_j \in \mathcal{C}} \exp \left(\mathbf{z}_{\mathbf{y}_j}^{r}/\sigma^{r}\right)},\ \ \hat{\mathbf{y}}^{c}_i=\frac{\exp \left(\mathbf{z}_{\mathbf{y}_i}^{c}/\sigma^{c}\right)}{\sum_{\mathbf{y}_j \in \mathcal{C}} \exp \left(\mathbf{z}_{\mathbf{y}_j}^{c}/\sigma^{c}\right)},
\end{equation}
where $\mathbf{z}^{r}$/$\mathbf{z}^{c}$ is the logits given by \textit{ref} $\mathcal{W}_{\phi}^{r}$/\textit{cls} $\mathcal{W}_{\phi}^{c}$ respectively and $\sigma$ is corresponding standard deviation for normalization. Then, we re-weight the KD loss for each sample as follows:
\begin{equation}
    \hat{w}_{\mathbf{x}_i} = \mathcal{H}(\hat{\mathbf{y}}_i^r, \mathbf{y}_i)/\mathcal{H}(\hat{\mathbf{y}}_i^c, \mathbf{y}_i) = \frac{\sum_{\mathbf{y}_j \in \mathcal{C}} \mathbbm{1}(\mathbf{y}_j=\mathbf{y}_i)\log \hat{\mathbf{y}}_j^{c}}{\sum_{\mathbf{y}_j \in \mathcal{C}} \mathbbm{1}(\mathbf{y}_j=\mathbf{y}_i)\log \hat{\mathbf{y}}_j^{r}},
    \label{eq:rtg_weight}
\end{equation}
where $\mathcal{H}$ is the corresponding cross-entropy. With Eq.~\ref{eq:rtg_weight}, we assign the under-represented samples (\textit{cls} head is less confident than the \textit{ref} head) with a large weight and the over-represented samples (\textit{cls} head is more confident than the \textit{ref} head) with a small weight.

\subsection{Balanced Online Distillation}
Following NCL~\cite{NCL}, we employ the online distillation framework to learn multiple experts collaboratively. On the logit level, we implement the re-weighted KD loss between each expert pair and the total loss will be:
\begin{equation}
    \mathcal{L}_{\text{kd}}^{\text{logit}}=\frac{\sum_{k} \sum_{q \neq k} \sum_{\mathbf{x}_i} \hat{w}_{\mathbf{x}_i} \cdot \tau^2 \cdot \text{KL}(\varsigma(\frac{\mathbf{z}_i^{k,c}}{\tau}) || \varsigma(\frac{\mathbf{z}_i^{q,c}}{\tau}))}{N\cdot K\cdot (K-1)}, \!
\end{equation}
where $\varsigma$ indicates \textit{softmax}, $\tau$ is the temperature factor and $\mathrm{KL}(p || q)=\sum_{i} p_{i} \cdot \log (p_{i}/q_{i})$. $\mathbf{z}_i^{k,c}$ is the logits given by the \textit{cls} head $\mathcal{W}_{\phi}^{c}$ of expert $k$ and $\hat{w}_{\mathbf{x}_i}$ is given by Eq.~\ref{eq:rtg_weight}.

Different from previous methods, we pinpoint that the distillation on the feature level will capture more robust knowledge in the LTR tasks, which can be formulated as follows:
\begin{equation}
    \mathcal{L}_{\text{kd}}^{\text{feature}} = \frac{\sum_k \sum_{q\neq k} \sum_{\mathbf{x}_i} \tau^2 \cdot \text{KL}(\varsigma(\frac{\mathbf{v}_i^{k}}{\tau}) || \varsigma(\frac{\mathbf{v}_i^{q}}{\tau}))}{K(K-1)}.
\end{equation}

Experimentally, the online distillation on feature level shows significant effectiveness compared to logit level. we will present in-depth investigations on the reason for its performance and generalization in Sec.~\ref{sec:exp}.

\subsection{Contrastive Proxy Task}
To learn more generalized features, we follow~\cite{MoCo} to adopt a contrastive proxy task in MoCo v2 manner. As Fig.~\ref{fig:single_pipeline} shows, an extra MoCo encoder is employed to perform instance discrimination, in which parameters are updated in a momentum-based moving average scheme to provide negative samples. For the feature $\mathbf{v}_i^k$ given by expert $k$ MoCo head, we denote the normalized embedding of its copy image with different augmentations as $\tilde{\mathbf{v}}_i^k$. For more negative pairs, a dynamic queue $\mathcal{Q}^{k}$ is employed to record historical feature representations to save GPU memory. The info-NCE loss is adopted to increase the feature similarity of the same image while reducing the feature similarity of different images pairs, which is computed as:
\begin{equation}
    \mathcal{L}_{\text{con}}= - \sum_{k} \log \frac{\exp \left(\mathbf{v}_{i}^{k^{T}} \tilde{\mathbf{v}}_{i}^{k}/\tau\right)}{\sum_{\tilde{\mathbf{v}}_{j}^{k} \in \{\mathcal{Q}^{k} \cup \mathbf{v}_{i}^{k^{T}} \} } \exp \left(\mathbf{v}_{i}^{k^{T}} \tilde{\mathbf{v}}_{j}^{k}/\tau\right)}.
\end{equation}

\begin{table*}[t!]
\caption{Top-1 accuracy (\%) on CIFAR-10/100-LT with ResNet32 backbone. $\gamma$: imbalance factor. Results are sorted according to method category. RW: re-weight wise methods. FW: feature improvement wise methods. ME: multi-expert frameworks. {\ul Underline}: the best performance in each group. {\textbf{Bold}}: the best performance overall. We report the performance from original papers and reproduce results for unavailable settings according to their official repos.}
\resizebox{1\linewidth}{!}{%
\centering
\setlength{\tabcolsep}{12pt}
\begin{tabular}{@{}l|ccc|cccc|cccc@{}}
\toprule
Dataset         & \multicolumn{3}{c|}{Type} & \multicolumn{4}{c|}{CIFAR100} & \multicolumn{4}{c}{CIFAR10}   \\ \midrule
$\gamma$              & RW      & FW     & ME     & 10    & 50    & 100   & 200   & 10    & 50    & 100   & 200   \\ \midrule
CE \cite{CB}                  & -       & -      & -      & 55.7 & 44.0 & 38.3 & 34.6 & 86.4 & 75.0 & 70.4 & 66.2 \\ \midrule
Focal Loss \cite{Focal}       & \checkmark       &        &        & 55.8 & 44.3 & 38.4 & 35.6 & 86.6 & 76.7 & 70.4 & 68.9 \\
$\tau$ Norm \cite{NCM}        & \checkmark       &        &        & 59.1 & 48.2 & 43.6 & 39.3 & 87.8 & 82.8 & 75.1 & 70.3 \\
Causal Norm \cite{CausalNorm} & \checkmark       &        &        & 59.6 & 50.3 & 44.1 & -     & {\ul 88.5} & 83.6 & {\ul 80.6} & -     \\
LADE \cite{LADE}              & \checkmark       &        &        & 61.6 & 50.1 & 45.6 & {\ul 40.7} & 88.3 & 82.1 & 79.1 & {\ul 73.9} \\
DRO \cite{DRO}                & \checkmark       &        &        & {\ul 63.4} & {\ul 57.6} & {\ul 47.3} & -     & -     & -     & -     & -     \\
TDE + IDR \cite{IDR}          & \checkmark       &        &        & -     & 50.3 & 44.9 & -     & -     & {\ul 84.5} & 79.6 & -     \\ \midrule
M2m \cite{M2m}                &         & \checkmark      &        & 58.2 & -     & 42.9 & -     & 87.9 & -     & 78.3 & -     \\
CAM \cite{Bagoftricks}        &         & \checkmark      &        & -     & 51.7 & 47.8 & -     & -     & {\ul 83.6} & {\ul 80.0} & -     \\
DiVE (2 Experts) \cite{DiVE}              &         & \checkmark      & \checkmark      & {\ul 62.0} & 51.1 & 45.4 & -     & -     & -     & -     & -     \\
CMO+RIDE (4 Experts) \cite{CMO}           &         & \checkmark      & \checkmark      & 60.2 & {\ul 53.0} & {\ul 50.0} & -     & -     & -     & -     & -     \\
TSC \cite{TSC}                &         & \checkmark      &        & 59.0 & 47.4 & 43.8 & -     & {\ul 88.7} & 82.9 & 79.7 & -     \\ \midrule
LDAM+DRW \cite{LDAM}          & \checkmark       & \checkmark      &        & 58.7 & 46.6 & 42.0 & 38.5 & 88.2 & 81.3 & 77.0 & 74.7 \\
MiSLAS \cite{MiSLAS}          & \checkmark       & \checkmark      &        & 63.2 & 52.3 & 47.0 & -     & 90.0 & 85.7 & 82.1 & -     \\
Prior-LT \cite{PriorLT}       & \checkmark       & \checkmark      &        & 61.3 & 51.1 & 45.5 & 42.1 & 89.7 & 84.3 & 82.8 & 78.5 \\
PaCo \cite{PaCo}              & \checkmark       & \checkmark      &        & 64.2 & 56.0 & {\ul 52.0} & {\ul 47.8} & {\ul 91.5} & {\ul 88.0} & {\ul 85.4} & {\ul 82.3} \\
BCL \cite{BCL}                & \checkmark       & \checkmark      &        & {\ul 64.9} & {\ul 56.6} & 51.9 & -     & 91.1 & 87.2 & 84.3 & -     \\
GCL \cite{GCL}                & \checkmark       & \checkmark      &        & -     & 53.6 & 48.7 & 44.9 & -     & 85.5 & 82.7 & 79.0 \\ \midrule
LFME (3 Experts) \cite{LFME}              &         &        & \checkmark      & 57.8 & 47.2 & 42.3 & 39.0 & 87.1 & 81.5 & 75.3 & 72.9 \\
BBN (2 Experts) \cite{BBN}                &         & \checkmark      & \checkmark      & 59.1 & 47.0 & 42.6 & -     & 88.3 & 82.2 & 79.8 & -     \\
RIDE (4 Experts) \cite{RIDE}   &         &        & \checkmark      & 61.8 & 51.7 & 48.0 & 44.6 & 86.3 & 83.7 & 81.2 & 77.8 \\
Hybrid-SC (2 Experts) \cite{HybirdSC}     &         & \checkmark      & \checkmark      & -     & 51.9 & 46.7 & -     & -     & 85.4 & 81.4 & -     \\
SADE (3 Experts) \cite{TADE}              & \checkmark       &        & \checkmark      & 63.6 & 53.9 & 49.8 & 44.7 & 90.0 & 85.8 & 82.9 & 78.0 \\
SSD (2 Experts) \cite{SSD}                &         &        & \checkmark      & 62.3 & 50.5 & 46.0 & -     & -     & -     & -     & -     \\
ACE (4 Experts) \cite{ACE}     &         &        & \checkmark      & -     & 51.9 & 49.6 & -     & -     & 84.9 & 81.4 & -     \\
NCL (3 Experts) \cite{NCL}     & \checkmark       &        & \checkmark      & {\ul 63.8} & {\ul 58.2} & {\ul 54.2} & {\ul 49.5} & {\ul 91.1} & {\ul 87.3} & {\ul 85.5} & {\ul 82.2} \\ \midrule
{\name} (3 Experts)            & \checkmark       & \checkmark      & \checkmark      & \textbf{67.3} & \textbf{59.9} & \textbf{56.3} & \textbf{51.4} & \textbf{91.8} & \textbf{88.9} & \textbf{86.5} & \textbf{83.6} \\
\bottomrule
\end{tabular}
}
\label{tab:cifar}
\end{table*}

\subsection{Model Training}
Based on the above designs, we propose our final {\name} in the multi-expert architecture, as Fig.~\ref{fig:teaser}\&\ref{fig:single_pipeline} shows. To train the whole model, we leverage the classification loss $\mathcal{L}_{\text{sup}}$, distillation loss $\mathcal{L}_{\text{kd}}$, and contrastive loss $\mathcal{L}_{\text{con}}$ for supervision. Formally, $\mathcal{L}_{\text{sup}}$ compute the all experts BC loss between the predicted logits and the ground-truth labels for \textit{ref} \& \textit{cls} head:
\begin{equation}
        \mathcal{L}_{\text{sup}} = \frac{1}{K} \sum_k \left( \mathcal{L}^{\text{ref}}_{\text{BC}} + \mathcal{L}^{\text{cls}}_{\text{BC}} \right )
\end{equation}

$\mathcal{L}_{\text{kd}}$ estimate the KL divergence on logit \& feature level, while $\mathcal{L}_{\text{con}}$ is used for the contrastive proxy task. Finally, the overall loss is formulated as:
\begin{equation}
    \mathcal{L}_{\text{all}} = \mathcal{L}_{\text{sup}} + \alpha (\mathcal{L}_{\text{kd}}^{\text{logit}} + \mathcal{L}_{\text{kd}}^{\text{feature}}) + \beta \mathcal{L}_{\text{con}},
    \label{eq:final_loss}
\end{equation}
where $\alpha$ and $\beta$ are the hyperparameters to balance the contribution of collaborative and contrastive learning.

\subsection{Model Inference}
Note that the MoCo branch and \textit{ref} head $\mathcal{W}_{\phi}^{r}$ are only designed for effective model training. Therefore, in the inference phase, we only preserve the feature encoder $\mathcal{F}_{\theta}$ and \textit{cls} head $\mathcal{W}_{\phi}$ to keep consistent model size with previous work. In addition, we can achieve higher performance by averaging the output logits from all experts as an ensemble model. In this case, our model size will be the same as the previous NCL.
\section{Experiment}
\label{sec:exp}

\subsection{Datasets}
\myparagraph{CIFAR-10/100-LT.} CIFAR-10/100~\cite{Cifar} have $10/100$ classes with $60,000$ images in $32\times32$ resolution. We follow~\cite{CB, LDAM} to sample the train set of each class with exponential functions to create the long-tailed versions while remaining the validation set uniformly distributed. The imbalance factor $\gamma$ indicates the skewness of the dataset, which is the ratio between the most and the least frequent classes. We employ $\gamma=[10,50,100,200]$ for comprehensive comparisons.

\myparagraph{ImageNet-LT} is the subset of the large-scale balanced ImageNet-1k~\cite{Imagenet}, widely used in classification and localization tasks. The train data in ImageNet-LT are sampled through Pareto distribution with power value $\alpha = 6$. It contains $115.8K$ images from $1,000$ classes. The most/least class number is $1,280/5$ respectively ($\gamma=256$). we utilize the balanced validation set constructed by~\cite{CB} for fair comparisons.

\myparagraph{iNaturalist 2018~\cite{iNaturalist}} is the large-scale real-world LTR dataset. With over $437.5K$ images and $8,142$ classes ($\gamma=500$), it suffers from severe label long-tailed distribution and fine-grained challenges. We follow~\cite{LDAM} to utilize the official splits of training and validation sets in our experiments.

\begin{table}[!t]
\centering
\caption{Top-1 accuracy (\%) on ImageNet-LT \& iNaturalist 2018. Results are sorted by publication time. R-50: ResNet-50. RX-50: ResNeXt-50. Our {\name} consistently outperforms state-of-the-art \czh{by a large margin}.}
\resizebox{1\linewidth}{!}{%
\centering
\setlength{\tabcolsep}{9pt}
\begin{tabular}{@{}l|cc|c@{}}
\toprule
\multicolumn{1}{c|}{\multirow{2}{*}{Method}} & \multicolumn{2}{c|}{ImageNet-LT} & iNaturalist2018 \\ \cmidrule(l){2-4} 
\multicolumn{1}{c|}{}                        & R-50            & RX-50          & R-50   \\ \midrule
CE \cite{CB}                                 & 38.9           & 44.4          & 60.9  \\
OLTR \cite{OLTR}                             & 40.4           & -              & 63.9  \\
CB \cite{CB}                                 & 40.9           & -              & 63.5  \\
LDAM+DRW \cite{LDAM}                         & 45.8           & -              & 68.0  \\
BBN \cite{BBN}                               & 48.3           & 49.3          & 66.3  \\
NCM \cite{NCM}                               & 44.3           & 47.3          & 63.1  \\
c-RT \cite{NCM}                              & 47.3           & 49.6          & 65.2  \\
$\tau$-Norm \cite{NCM}                       & 46.7           & 49.4          & 65.6  \\
LWS \cite{NCM}                               & 47.7           & 49.7          & 65.9  \\
BS \cite{BS}                                 & 53.0           & -              & 66.4  \\
RIDE (4 Expert) \cite{RIDE}                  & 55.4          & 56.8          & 72.6  \\
DisAlign \cite{DisAlign}                     & 52.9           & 53.4          & 70.6  \\
DiVE \cite{DiVE}                             & 53.1           & -              & 71.7  \\
SSD (2 Expert) \cite{SSD}                    & -               & 56.0          & 71.5  \\
ACE (4 Expert)  \cite{ACE}                   & 54.7           & 56.6          & 72.9  \\
PaCo \cite{PaCo}                             & 56.1           & 57.2          & 72.2  \\
TSC  \cite{TSC}                              & 52.4           & -              & 69.7  \\
RIDE+CMO (4 Expert) \cite{CMO}               & 56.2           & -              & 72.8  \\
BCL \cite{BCL}                               & 56.0           & 57.1          & 71.8  \\
CKT \cite{CKT}                               & -               & 54.2          & -      \\
GCL \cite{GCL}                               & 53.7           & 54.9          & 72.0  \\
NCL (3 Expert) \cite{NCL}                    & 59.5           & 60.5          & 74.9  \\
{\name} (3 Expert)                               & \textbf{60.6}           & \textbf{61.7}          & \textbf{75.8}  \\ \bottomrule
\end{tabular}
}
\label{tab:large}
\end{table}

\subsection{Evaluation Metrics}
\label{sec:metrics}
\myparagraph{Top-1 Acc.}
In LTR, the model is trained in an imbalanced dataset while evaluated in a balanced test set. Therefore, we adopt the common evaluation protocol Top-1 Acc. to estimate the model performance of each category.

\myparagraph{Group Acc.} 
In LTR, we focus more on the tail performance. Therefore, we follow~\cite{TADE} to group the test data into Many-shot ($>100$), Medium-shot ($20\sim100$), and Few-shot ($< 20$) according to the corresponding sample number w.r.t. different classes in the train set, and evaluate the accuracy of these groups separately.

\myparagraph{Loss-Acc Landscapes.}
To investigate the feature-level distillation, we visualize the loss/accuracy landscapes of different models~\cite{Landscape}. More specifically, we perturb the model weights by varying degrees through a series of Gaussian noises. The noise level is normalized to the $l_2$-norm of each filter to represent the effects of different weight amplitudes.

\begin{table}[t]
\centering
\caption{Ablation study of {\name}. We report ResNet32 on CIFAR100-LT ($\gamma=100$) and ResNet50 on ImageNet-LT. BKT: balanced knowledge transfer module. FLD: feature level distillation. CPT: contrastive proxy task.}
\resizebox{1\linewidth}{!}{%
\begin{tabular}{@{}ccc|cc|cc@{}}
\toprule
BKT & FLD & CPT & CIFAR100-LT & $\Delta$   & ImageNet-LT & $\Delta$   \\ \midrule
-   & -   & -                               & 52.2         & -   & 55.1        & -   \\
\checkmark & -   & -                        & 53.7         & + 1.5 & 56.4        & + 1.3 \\
-    & \checkmark  & -                      & 54.7         & + 2.5 & 58.9        & + 3.8 \\
-   & -   & \checkmark                      & 54.2         & + 2.0 & 57.6        & + 2.5 \\
\checkmark   & -    & \checkmark            & 54.9         & + 2.7 & 57.9        & + 2.8 \\
-    & \checkmark   & \checkmark            & 55.8         & + 3.6 & 59.9        & + 4.8 \\
\checkmark   & \checkmark   & \checkmark    & 56.3         & + 4.1 & 60.6        & + 5.5 \\ \bottomrule
\end{tabular}
}
\label{tab:ablation}
\end{table}

\myparagraph{Class-wise Average Feature Distance.}
On the balanced test dataset, a well-trained encoder should map the input images into a distinguishable feature space. To evaluate the feature similarity, For class $\mathbf{y}_i$, we calculate the class-wise average $l_2$ distance between the outputs from two experts for all features ($\mathcal{A}_{\mathbf{y}_i}$). Here, we calculate the distance between expert $m$ and $n$ as follows: 
\begin{equation}
    D_{i}^{m,n} = \frac{1}{||\mathcal{A}_{\mathbf{y}_i}||} \cdot \sum_{\mathbf{v}_t \in \mathcal{A}_{\mathbf{y}_i}} ||\mathbf{v}_t^m - \mathbf{v}_t^n||_2.
\end{equation}

The smaller $D_i$ indicates that the experts learn stable feature representations w.r.t. class $\mathbf{y}_i$, making it easier to finetune model heads on the downstream tasks.

\myparagraph{Expected Calibration Error.}
Calibration indicates the model prediction reflects the actual likelihood of accuracy~\cite{ECE}. Let $\hat{p}_{i}$ be the confidence of the image $x_i$, and divide dataset $\mathcal{D}$ into several bin $\mathcal{B}$ with size $m$ according to the value of $\hat{p}_{i}$. Then, the reliability diagrams are proposed to visualize the model calibration by measuring the distance to the ideal $\sum_{i \in \mathcal{B}_m} \mathbbm{1}(\hat{y}=y_i) \equiv \sum_{i \in \mathcal{B}_m} \mathbbm{1}(\hat{y}=y_i)$ for all $m\in \{1,..., M\}$. The Expected Calibration Error (ECE) is proposed to quantitatively measure classifiers' calibration:
\begin{equation}
    ECE =  \frac{1}{|\mathcal{D}|} \sum_{m=1}^{M} \sum_{i \in \mathcal{B}_m} | \mathbbm{1}(\hat{y}=y_i) - \hat{p}_{i}|.
\end{equation}

\subsection{Implementation Details}
\label{sec:implement}
For CIFAR-LT, we follow LTR-WD~\cite{LTR-WD} to set weight decay $5e-3$ for ResNet-32 and use stochastic gradient descent with momentum $0.9$. All models are trained for $200$ epochs with the learning rate $0.01$ and mini-batch $64$. The learning scheduler is Cosine Annealing~\cite{CosLearning} with an ending rate of $0$. Further, Cutout~\cite{Cutout} and AutoAug~\cite{Autoaug} are used to compensate for origin data augmentation strategies~\cite{ResNet}. We adopt the MoCo augmentation~\cite{MoCo} for better image views in the contrastive branch. For large-scale datasets, we follow LTR-WD~\cite{LTR-WD} to set weight decay $5e-4$/$1e-4$ for ImageNet-LT/iNaturalist 2018 and train $180$/$90$ epochs, respectively. We replace AutoAug with RandAug~\cite{Randaugment} while keeping other settings consistent with CIFAR-LT. Finally, we adopt horizontal flips as the post-hoc augmentation for better performance. 

Following previous work~\cite{MoCo, PaCo, NCL}, we set temperature factor $\tau = 1$ and keep all MoCo hyper-parameters consistent with NCL. For the hyper-parameters setting of {\name}, we set $K=3$ experts, $\alpha=0.6$ and $\beta=1.0$ by default. Results are averaged from 5 (CIFAR-LT) or 3 (large-scale datasets) random seeds.

\begin{figure}[t]
	\centering
    \subfloat[Top-1 Acc]{
    	\begin{overpic}[trim=0cm 0cm 0cm 1cm,clip,width=0.47\linewidth,grid=False]{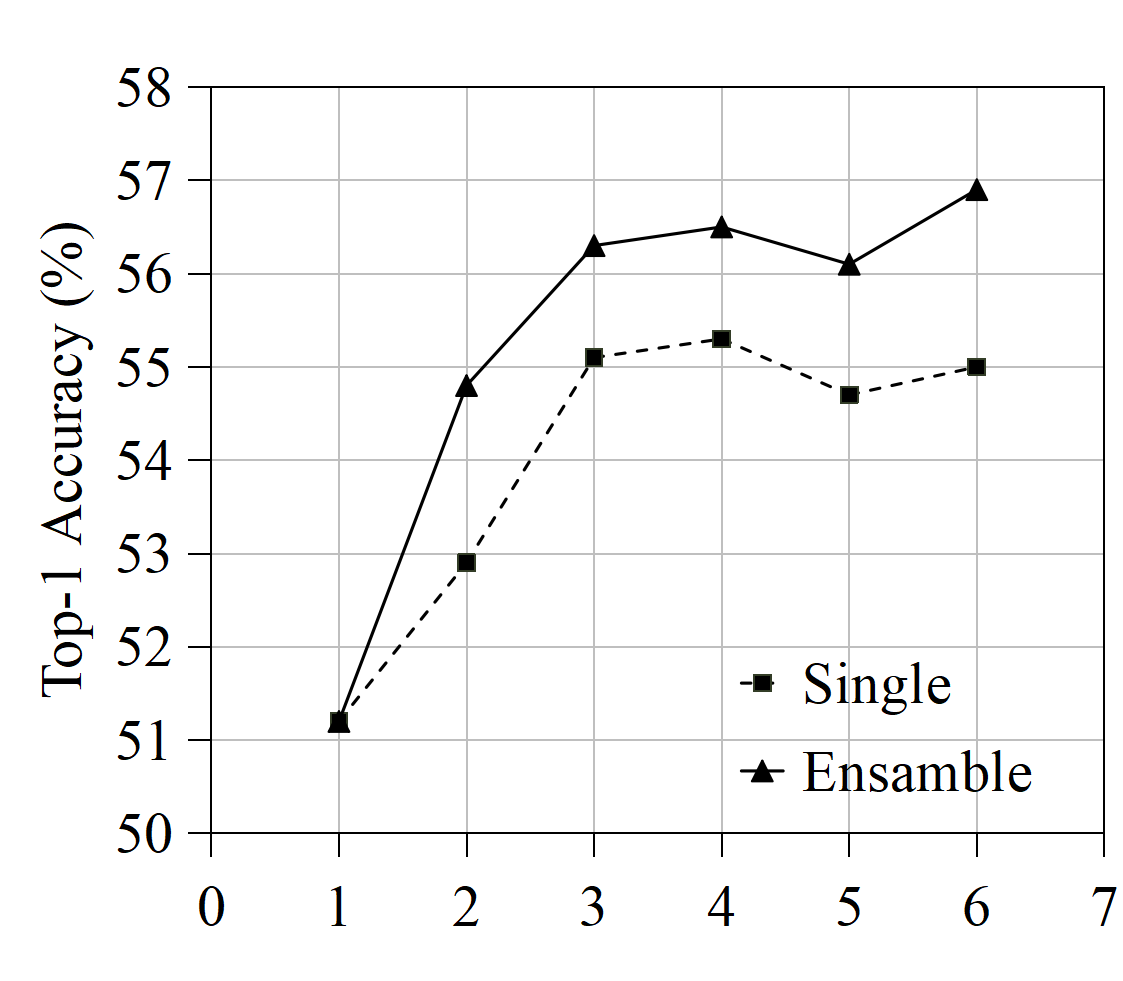}
        \end{overpic}
    }
    \subfloat[Few Acc]{
    	\begin{overpic}[trim=0cm 0cm 0cm 1cm,clip,width=0.47\linewidth,grid=False]{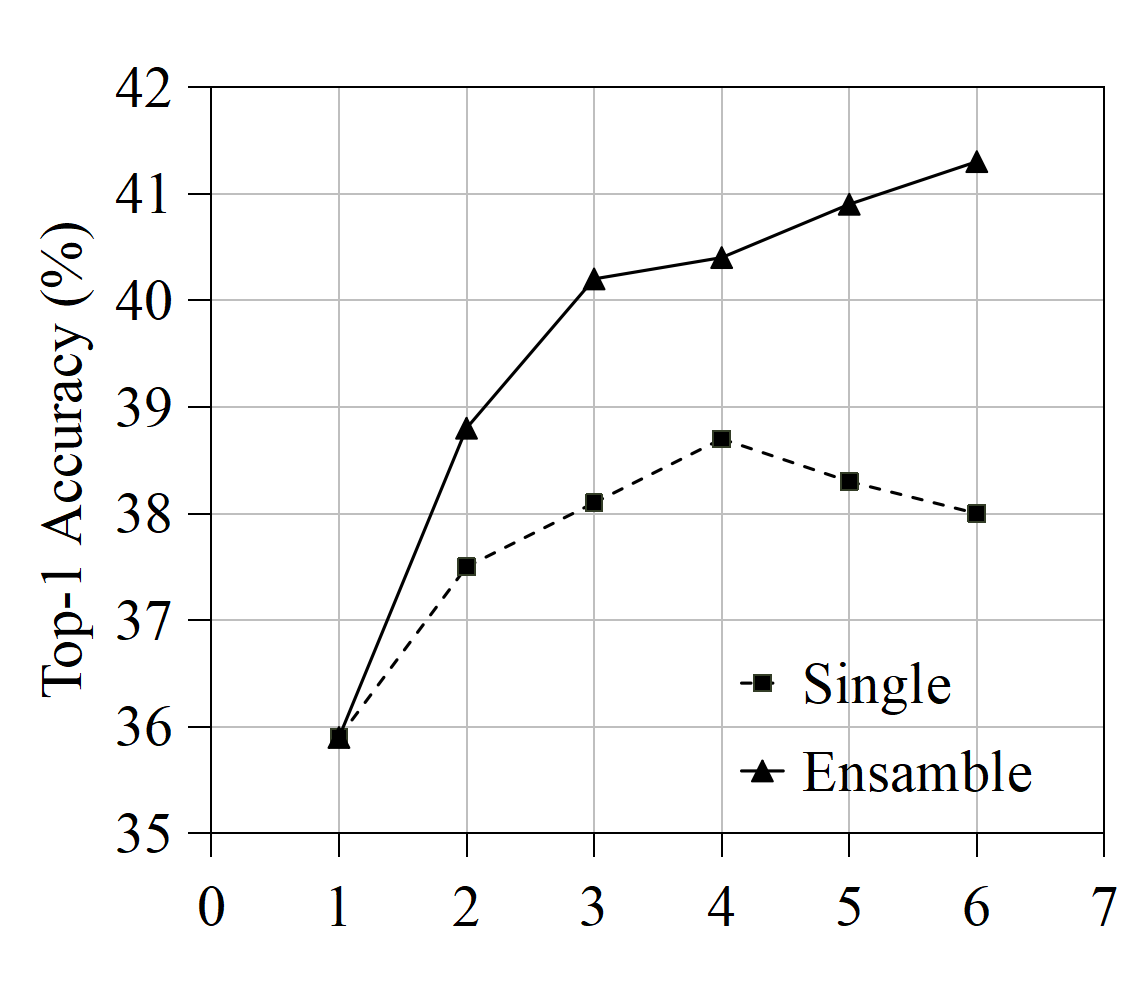}
        \end{overpic}    
    }
	\caption{Comparison of different expert number $K$ on CIFAT100-LT ($\gamma=100$). The ensemble performance is computed based on the averaging logits of all experts. We report Top-1/Few-shot Acc and show that a larger expert number $K$ brings higher model performance. We set $K=3$ to leverage the performance and training memory consumption.}
    \label{fig:expert_number}
	\vspace{-15pt}
\end{figure}

\subsection{Competing Methods}
\myparagraph{Baselines.} The vanilla baseline (CE) conducts plain training with standard cross-entropy loss~\cite{CB}. The common networks are ResNet-32 (CIFAR-10/100-LT), ResNet-50~\cite{ResNet} (ImageNet-LT, iNaturalist 2018) and ResNeXt-50~\cite{ResNeXt} (ImageNet-LT). In addition, to align with previous works that contain some additional proposal-independent tricks implicitly, we adopt the same settings with NCL for all our reproduced results for fair comparisons.

\myparagraph{Feature-wise methods} modify the feature sampling or learning manners to cope with long-tailed datasets. M2m~\cite{M2m} generates pseudo samples for training and optimizing. CAM~\cite{Bagoftricks} and CMO~\cite{CMO} enrich the training samples via feature combination. DiVE~\cite{DiVE} adopts knowledge distillation and takes the teacher feature as an additional training sample for the student model. Recent state-of-the-art~\cite{TSC, PaCo, BCL} adopts contrastive frameworks to improve representation learning.

\myparagraph{Re-weight methods} focus on label weighting~\cite{CB, Focal, LDAM, MiSLAS} or logits adjusting~\cite{LA, LADE, BS, PriorLT, DRO, IDR, GCL} based on standard cross entropy loss. In addition, some methods~\cite{NCM, CausalNorm, LTR-WD} are also effective by directly adjusting the classifier's weight.

\begin{figure}[t]
	\centering
    \subfloat[$\alpha$]{
    	\begin{overpic}[trim=0cm 0cm 0cm 1cm,clip,width=0.47\linewidth,grid=False]{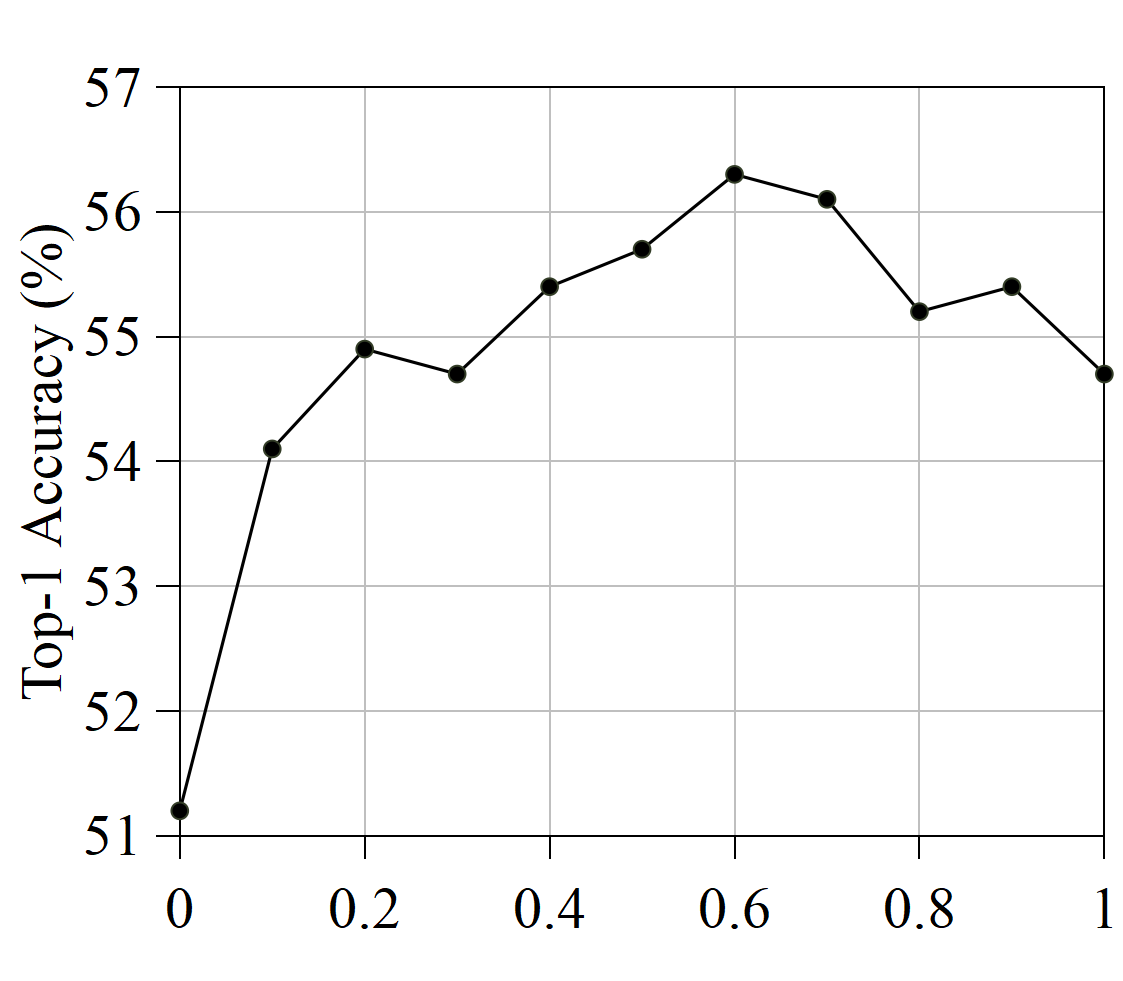}
        \end{overpic}
        \label{fig:alpha}
    }
    \subfloat[$\beta$]{
    	\begin{overpic}[trim=0cm 0cm 0cm 1cm,clip,width=0.47\linewidth,grid=False]{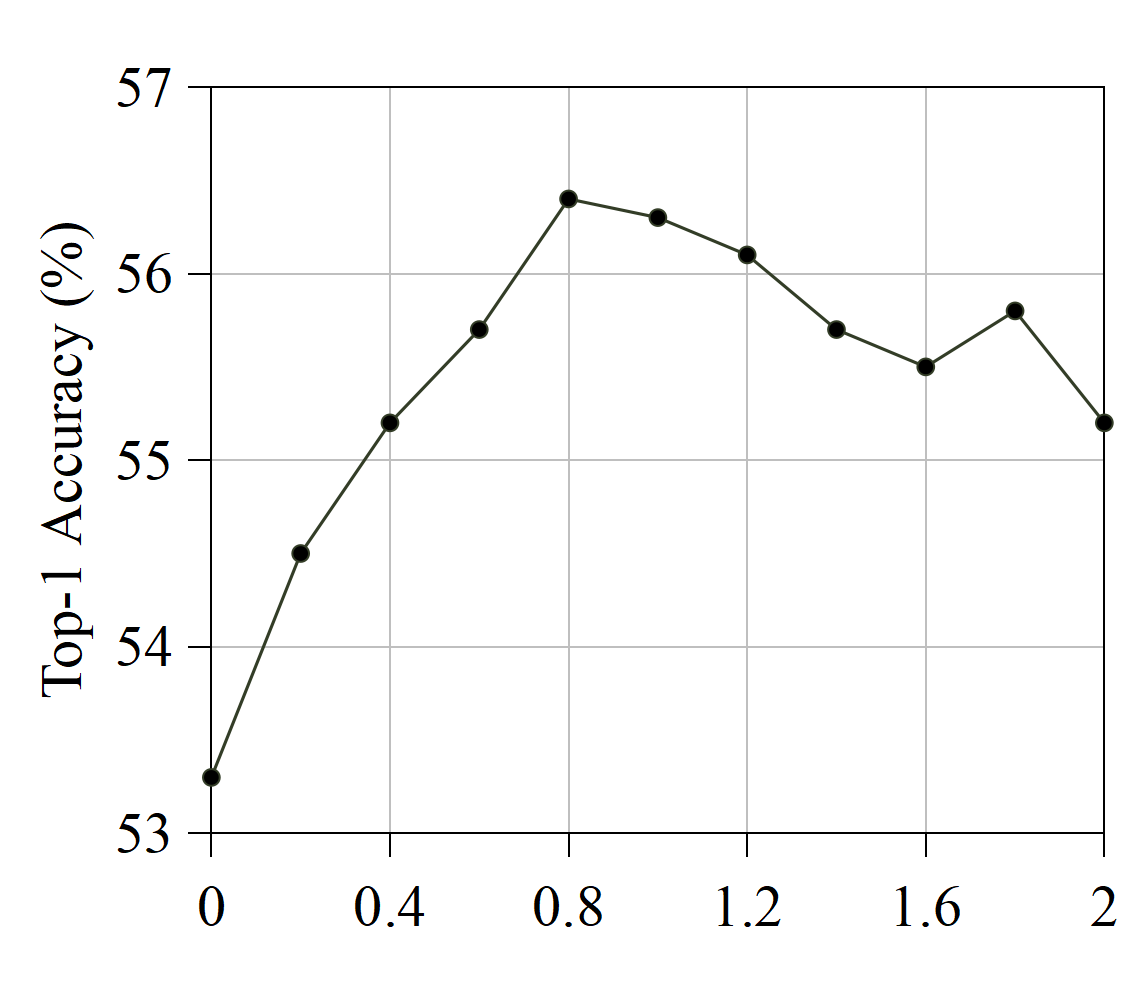}
        \end{overpic}
        \label{fig:beta}
    }
	\caption{Hyper-parameter analysis of $\alpha$ and $\beta$ on CIFAR100-LT ($\gamma=100$). We fix $\beta=1$ in subfigure (a) and $\alpha=0.6$ in subfigure (b). $\alpha=0$ means no collaborative learning in our {\name}, which results in poor Top-1 Acc performance.}
    \label{fig:hyper_para}
	\vspace{-15pt}
\end{figure}

\begin{figure}[!t]
	\centering
	\resizebox{\linewidth}{85pt}{
        \subfloat[CE\cite{CB}]{
            \includegraphics[width=0.3\linewidth, frame]
            {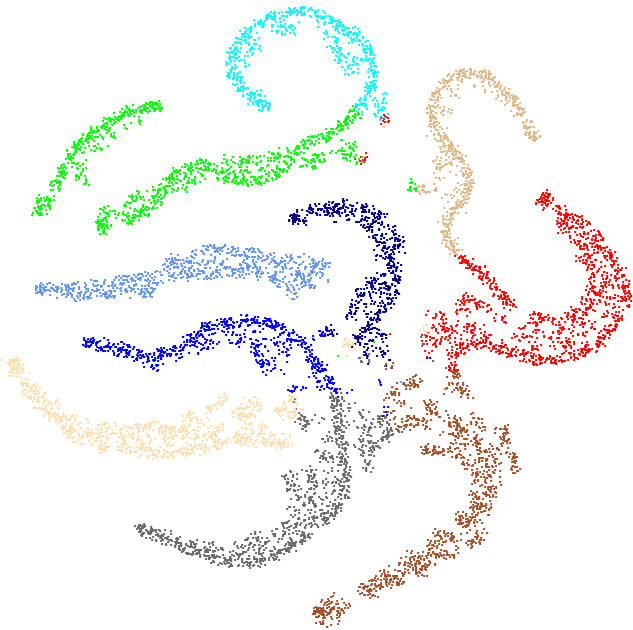}
            \label{fig:tsne_CE_10}
        }
        \subfloat[NCL\cite{NCL}]{
            \includegraphics[width=0.3\linewidth, frame]
            {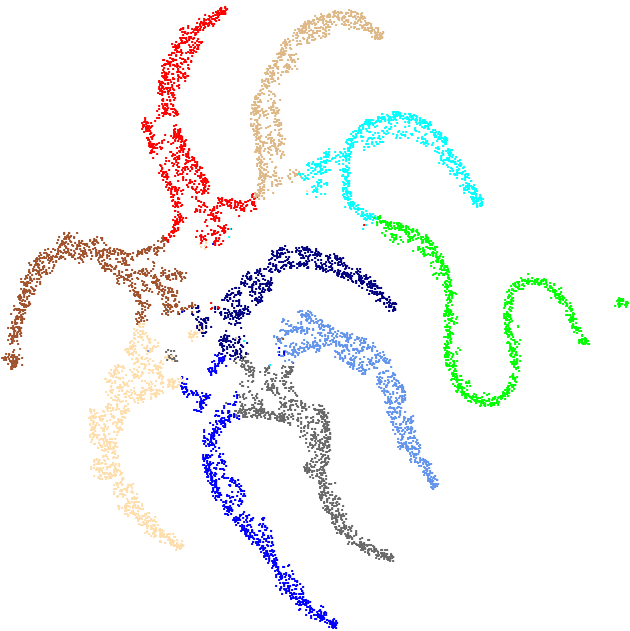}
            \label{fig:tsne_NCL_10}
        }
        \subfloat[{\name}]{
            \includegraphics[width=0.3\linewidth, frame]
            {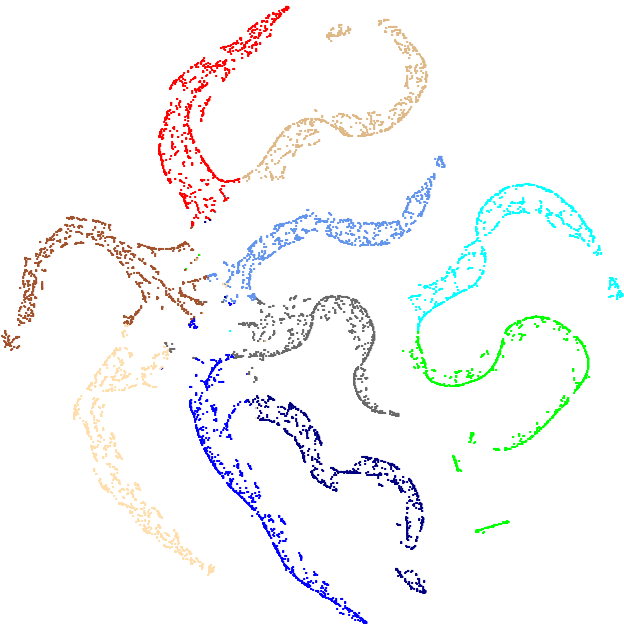}
            \label{fig:tsne_ECL_10}
        }
    }
    \\
    \resizebox{\linewidth}{85pt}{
        \subfloat[CE\cite{CB}]{
            \includegraphics[width=0.3\linewidth, frame]
            {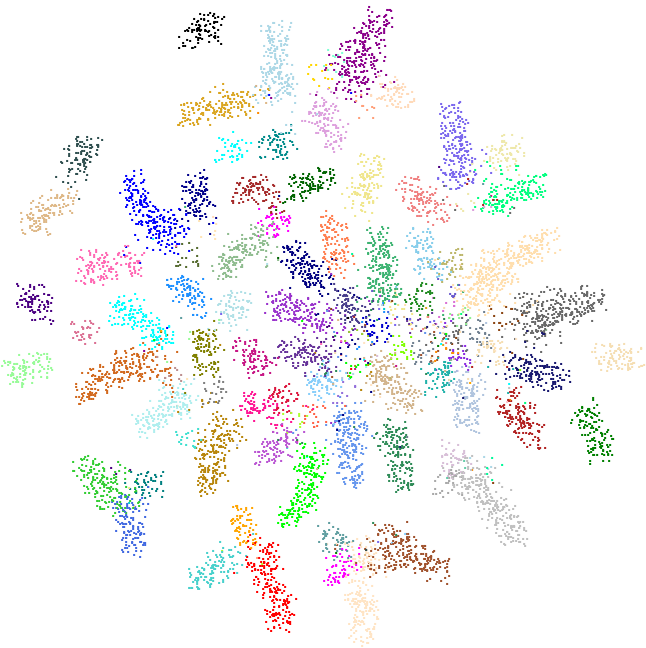}
            \label{fig:tsne_CE_100}
        }
        \subfloat[NCL\cite{NCL}]{
            \includegraphics[width=0.3\linewidth, frame]
            {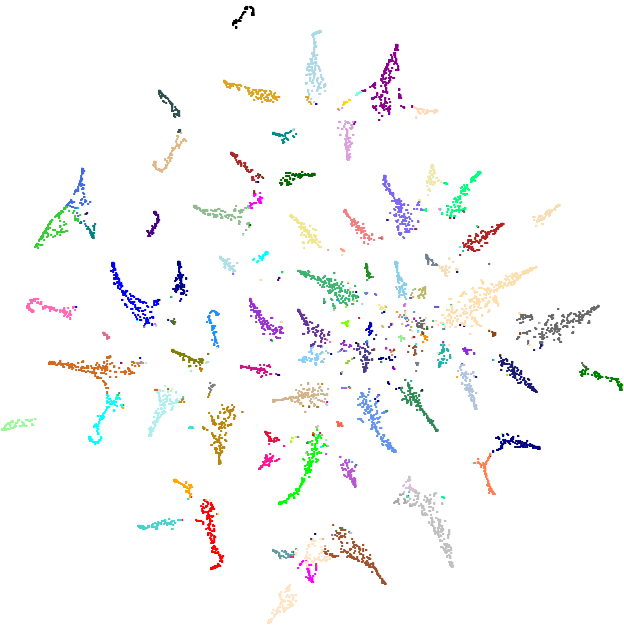}
            \label{fig:tsne_NCL_100}
        }
        \subfloat[{\name}]{
            \includegraphics[width=0.3\linewidth, frame]
            {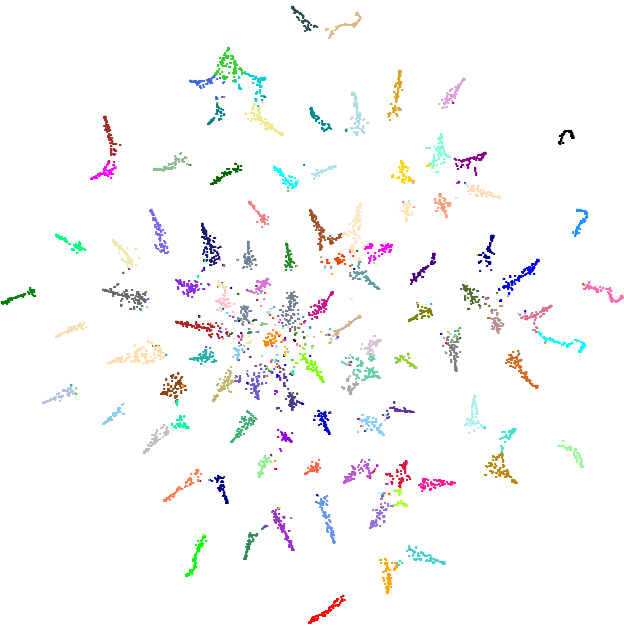}
            \label{fig:tsne_ECL_100}
        }
    }
	\caption{Visualized t-SNE results of ResNet32 on CIFAR10-LT (a-c) and CIFAR100-LT (d-e). The scatters of the same color indicate the same categories. Our {\name} shows better intra-class and inter-class distance to disentangle different categories.}
    \label{fig:tsne}
	\vspace{-15pt}
\end{figure}

\myparagraph{Multi-expert methods} have shown powerful generalization in LTR and can be classified into two categories. 1) Each expert learn \textit{different} aspects of knowledge w.r.t. specific classes and then aggregates together~\cite{LFME, BBN, RIDE, ACE, HybirdSC}. 2) Each expert learns the \textit{same} knowledge w.r.t. class to reduces the uncertainty on minority classes~\cite{TADE, SSD, NCL}. Note that our {\name} belongs to the latter.

\subsection{Comparison with state-of-the-art}
We conduct comprehensive comparisons on CIFAR-LT (see Tab.~\ref{tab:cifar}) and large-scale datasets (see Tab.~\ref{tab:large}). For comparing methods, we report the performance in their original papers and reproduce the missing settings through their official code repositories. For contrastive approaches~\cite{PaCo, NCL}, we keep the training epochs \textit{consistent} with ours for fair comparisons. We group previous methods into 3 categories as discussed in Sec.~\ref{sec:relate}. {\name} adopts RW (BC loss), FW (feature distillation), and ME (multi-expert architecture). Note that we report the ensemble results for all ME methods.

As illustrated in Tab.~\ref{tab:cifar}-\ref{tab:large}, {\name} outperforms previous approaches remarkably on all CIFAR-LT settings, ImageNet-LT, and iNaturalist 2018. Compared to state-of-the-art performance, {\name} improves the NCL by 2.1\% (CIFAR100-LT, $\gamma=100$), 1.1\% (ImageNet-LT), and 0.9\% (iNaturalist 2018) respectively. Compared to two-stage methods like MiSLAS~\cite{MiSLAS} and GCL~\cite{GCL}, our {\name} outperforms them in an end-to-end manner. Although we train the model in the multi-expert framework, we can adopt a single expert for evaluation without extra computation and memory consumption. We will discuss the single expert performance in Sec.~\ref{sec:further_analysis}.

\subsection{Further Analysis}
\label{sec:further_analysis}
\myparagraph{Ablation study of {\name}.}
We elaborately design three main modules to compound {\name}, namely the Balanced Knowledge Transfer module (BKT), Feature Level Distillation (FLD), and Contrastive Proxy Task (CPT). We conduct extensive ablation experiments on CIFAR100-LT ($\gamma=100$) to demonstrate the contribution of each component. As Tab.~\ref{tab:ablation} shows, our proposals are complementary to the performance, and FLD contributes the primary parts. As discussed in Sec.~\ref{sec:motivation} (observation \textbf{2}), FLD promotes more robust feature learning without the toxicity from prior label bias.
Like NCL~\cite{NCL} and PaCo~\cite{PaCo}, the CPT consistently improves model performance without inference burden. In addition, the BKT module consistently improves logit level distillation, allowing machine domain knowledge to be transferred with unbiased weight.

\myparagraph{Effect of expert number $K$.}
We conducted experiments to explore the influence of expert number $K$. As Fig.~\ref{fig:expert_number} shows, the model performance improves consistently with larger $K$. When $K=1$, the model is equal to the baseline with CPT without collaborative learning. When $K=2$, the performance improves significantly, which firmly manifests the effectiveness of BKT and FLD. However, when $K \geq 3$, the single expert is difficult to get further improvement, especially on few-shot accuracy. Hence, we set $K=3$ to trade off the computational overhead and model performance.

\myparagraph{Hyper-parameters analysis.}
In the final loss (Eq.~\ref{eq:final_loss}), we trade off the collaborative learning with $\alpha$ and contrastive learning with $\beta$. Fig.~\ref{fig:hyper_para} is designed to search for the optimal value on CIFAR100-LT ($\gamma=100$). In Fig.~\ref{fig:alpha}, we set $\beta=1$ by default. When $\alpha=0$, the model is degraded to the baseline with CPT. Top-1 Acc. increases rapidly when we add distillation loss ($\alpha>0$). The best trade-off between distillation and classification loss achieves at $\alpha=0.6$. In Fig.~\ref{fig:beta}, we set $\alpha=0.6$ by default. The best performance is achieved when $\beta \sim 1$, which shows a balance between classification and instance discrimination.

\begin{table}[t!]
\centering
\caption{Performance comparison with NCL in detail.}
\resizebox{\linewidth}{38pt}{
\begin{tabular}{@{}lc|ll|ll@{}}
\toprule
\multicolumn{2}{c|}{Dataset}                         & \multicolumn{2}{c|}{CIFAR100-LT} & \multicolumn{2}{c}{ImageNet-LT} \\ \midrule
\multicolumn{2}{c|}{Metric}                          & Acc $\uparrow$            & ECE $\downarrow$             & Acc $\uparrow$            & ECE $\downarrow$             \\ \midrule
\multicolumn{1}{l|}{NCL~\cite{NCL}} & \multirow{2}{*}{Single}   & 53.6           & 5.11            & 57.7           & 3.80            \\
\multicolumn{1}{l|}{ECL} &                           & 55.1 \good{(+1.5)}    & 2.33 \good{(-2.78)}    & 59.3 \good{(+1.6)}    & 1.96 \good{(-1.84)}   \\ \midrule
\multicolumn{1}{l|}{NCL~\cite{NCL}} & \multirow{2}{*}{Ensamble} & 54.4           & 4.62            & 59.5           & 2.92           \\
\multicolumn{1}{l|}{ECL} &                           & 56.3 \good{(+1.9)}    & 1.82 \good{(-2.80)}    & 60.6 \good{(+1.1)}    & 1.33 \good{(-1.59)}   \\ \bottomrule
\end{tabular}}
\vspace{-10pt}
\label{tab:cmp_ncl}
\end{table}

\begin{figure}[t!]
	\centering
	\begin{overpic}[trim=0cm 0cm 0cm 0cm,clip,width=1\linewidth,grid=False]{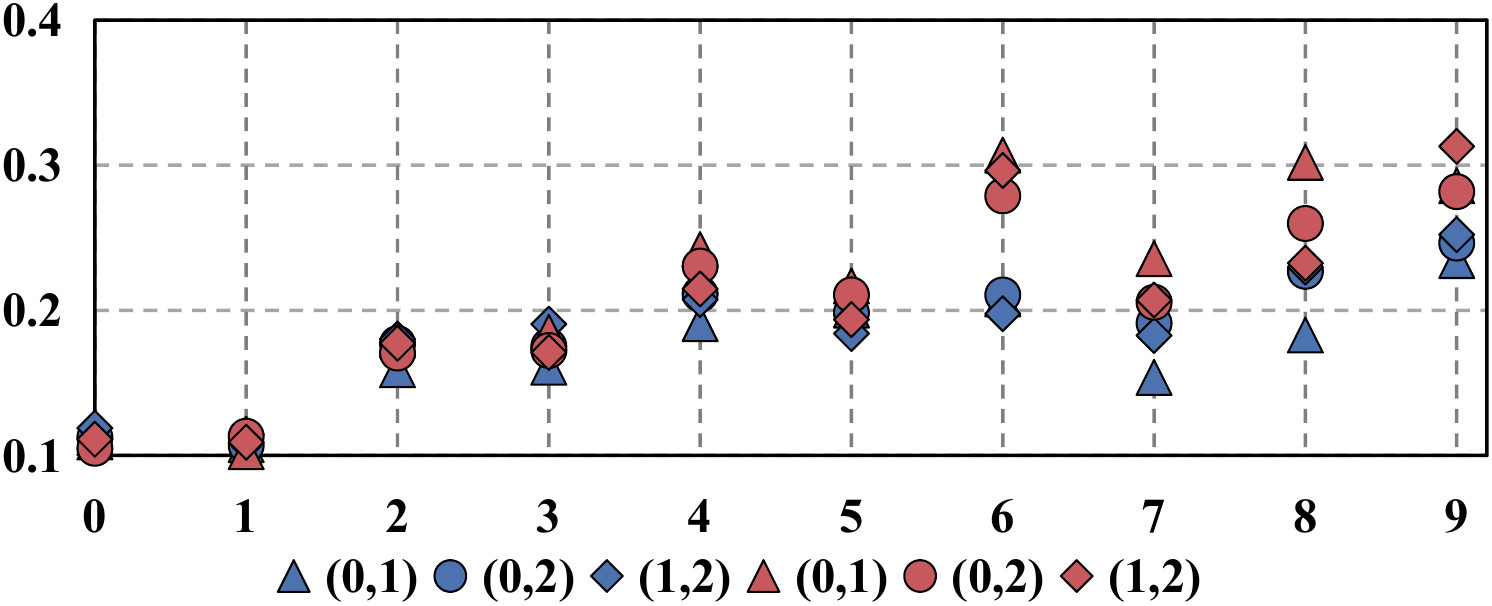}
    \end{overpic}
	\caption{The class-wise feature $l_2$ distance between each expert pair on CIFAR10-LT. The \textcolor{red}{red}/\textcolor{blue}{blue} points indicate NCL and {\name}. {\name} effectively reduces the differences between different experts on the feature of the same images.}
	\vspace{-15pt}
	\label{fig:fd_expert}
\end{figure}

\begin{figure}[t!]
	\centering
    \subfloat[w/o FLD]{
        \includegraphics[width=0.45\linewidth]
        {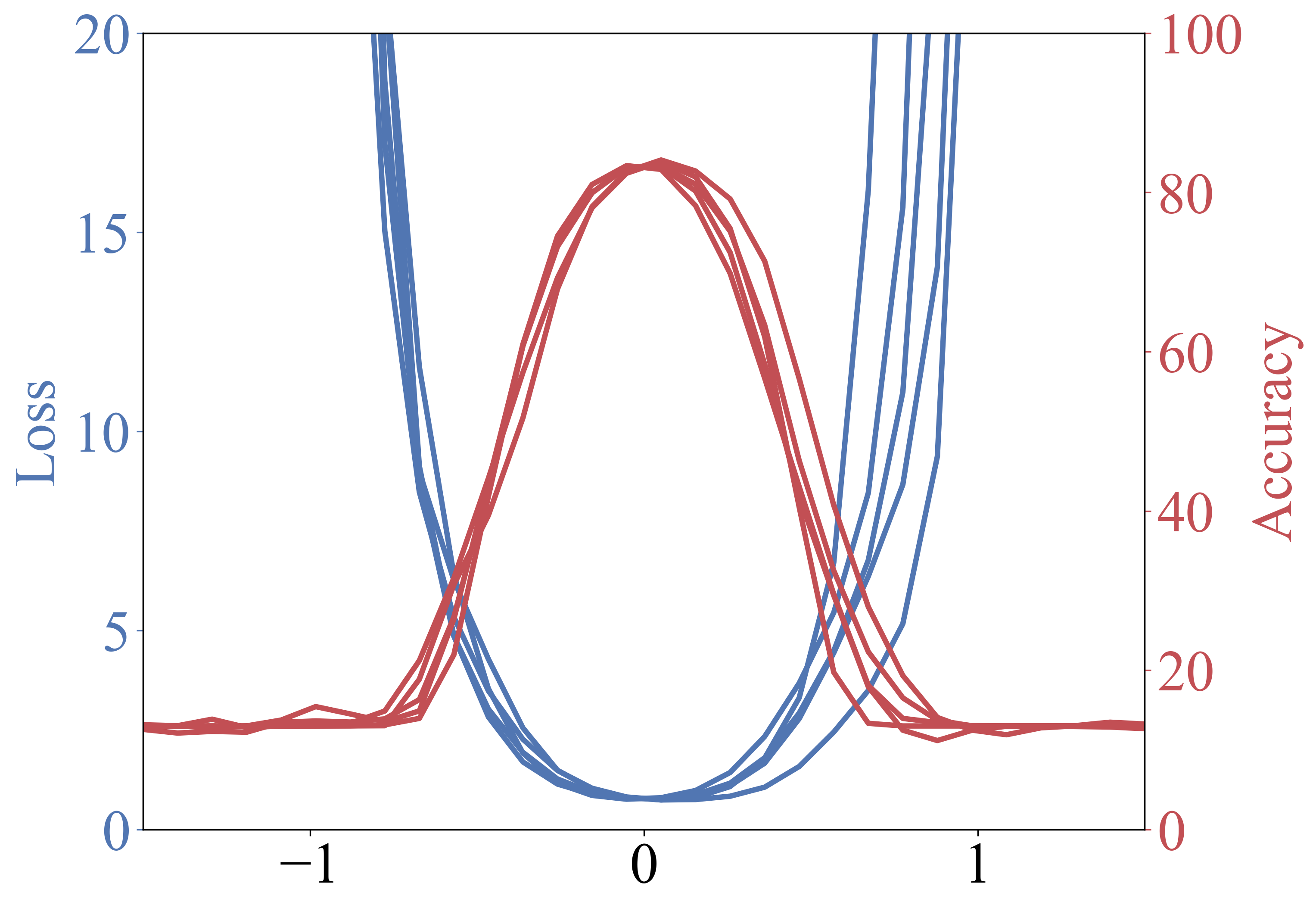}
    }
    \subfloat[w/i FLD]{
        \includegraphics[width=0.45\linewidth]
        {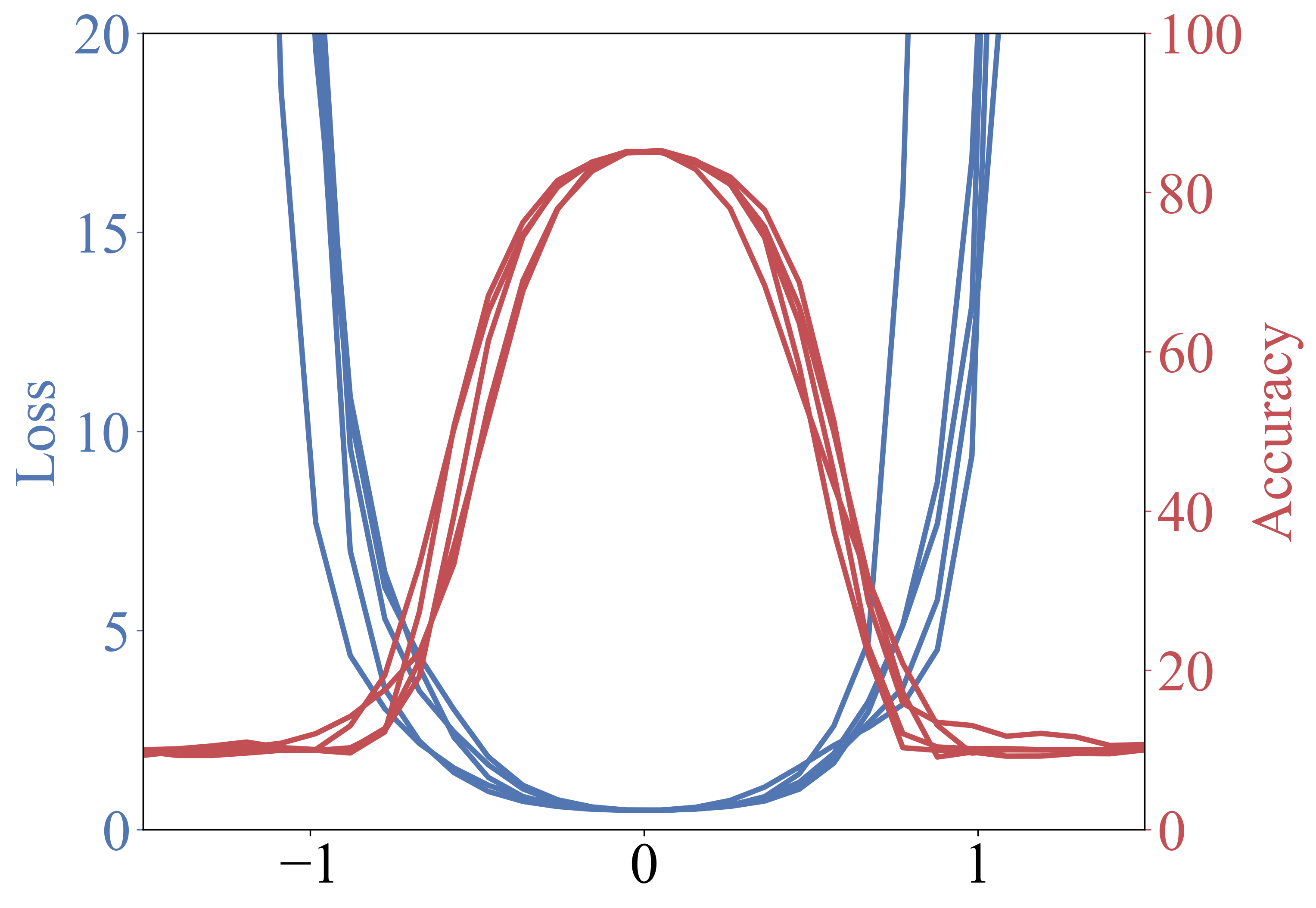}
    }
	\caption{The loss/accuracy landscapes of {\name} without (a) or with (b) feature level distillation. All plots contain $5$ landscapes with $5$ randomly generated directions.}
    \label{fig:landscape}
	\vspace{-15pt}
\end{figure}

\myparagraph{Feature representation quality.}
In Tab.~\ref{tab:ablation}, we notice that the feature level distillation is crucial in {\name}. To delve into its mechanism, we conduct visualization experiments on CIFAR10-LT and CIFAR100-LT ($\gamma=100$) in Fig.~\ref{fig:tsne}. Specifically, we utilize t-SNE\cite{tsne} to map the $K$-dimensional features into 2D distribution for visualization. Fig.~\ref{fig:tsne_CE_10}  \& \ref{fig:tsne_CE_100} show that the baseline cannot achieve satisfactory clustering results where few-shot categories are coupled together. The poor inter-class distance prevents further performance gains of the classifier. Note that collaborative learning (NCL) remarkably alleviates this issue as shown in Fig.~\ref{fig:tsne_NCL_10} \& \ref{fig:tsne_NCL_100}. Our {\name} further contributes to more compact intra-class distributions and enlarges the inter-class distance (Fig.~\ref{fig:tsne_ECL_10} \& \ref{fig:tsne_ECL_100}), which demonstrates that {\name} provides higher quality features.

In addition, we visualize the feature distance among each expert and summarize the average distance w.r.t. class index, which is sorted by instances number. As Fig.~\ref{fig:fd_expert} shows, all experts extract similar features of many-shot samples. However, the feature representations present differentiated distribution in few-shot samples. Our {\name} alleviates this problem remarkably via feature-level distillation, yielding its better classification performance.

\myparagraph{Loss/Accuracy landscapes.}
To validate the model robustness, we adopt the tool in~\cite{Landscape} to visualize the loss/accuracy landscapes of models with/without feature level distillation. We conduct experiments on CIFAR10-LT ($\gamma=100$) based on our {\name}. As described in Sec.~\ref{sec:metrics}, we perturb the model weights by a series of Gaussian noises with varying degrees. As Fig.~\ref{fig:landscape} shows, it turns out that the loss/accuracy landscapes become much flatter if we adopt the feature level distillation on {\name}. This observation demonstrates that the distillation operation help models to extract more robust representations to overcome the random noise perturbation.

\begin{figure*}[t!]
	\centering
	\begin{overpic}[trim=0cm 0cm 0cm 0cm,clip,width=1\linewidth,grid=False]{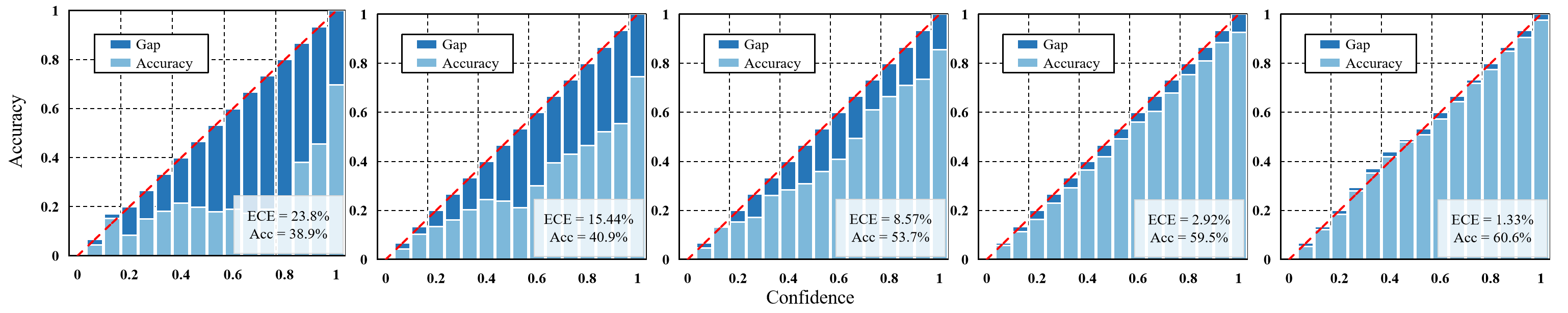}
	\put(6,12){CE}
	\put(25,12){CB~\cite{CB}}
	\put(44,12){GCL~\cite{GCL}}
	\put(63,12){NCL~\cite{NCL}}
	\put(83,12){{\name}}
    \end{overpic}
	\caption{Reliability diagrams on ImageNet-LT with $15$ bins. We select ResNet50 models trained via plain CE, CB, GCL, NCL, and our {\name}. The prediction probabilities of our {\name} indicate optimal expected costs in Bayesian decision scenarios.}
	\label{fig:reliablity}
	\vspace{-15pt}
\end{figure*}

\myparagraph{Model calibration.}
In Fig.~\ref{fig:reliablity}, we present the reliability diagrams with $15$ bins on the ImageNet-LT. For all models in comparison, the accuracy bars are below the ideal $y=x$ red line, which indicates that the models are all over-confident in their predictions. Compared to baseline CE, all methods alleviate the overconfidence issue and promote model calibration to some extent. Compared to NCL, {\name} further reduces ECE, which demonstrates our success in regulating all classes. We present more detailed comparisons to the state-of-the-art on the best single model and ensembles in Tab.~\ref{tab:cmp_ncl}. Our {\name} consistently outperforms the NCL in either single and ensemble views, and the single expert of {\name} achieves comparable performance with the NCL ensemble.

\begin{figure*}[ht!]
	\centering
	\subfloat[CE\cite{CB}]{
        \includegraphics[width=0.15\linewidth]
        {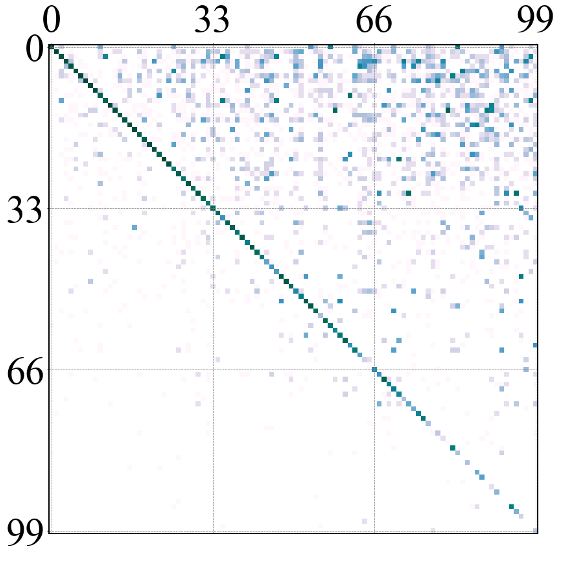}
        \label{fig:vis_cf_CE}
    }
    \subfloat[CB\cite{CB}]{
        \includegraphics[width=0.15\linewidth]
        {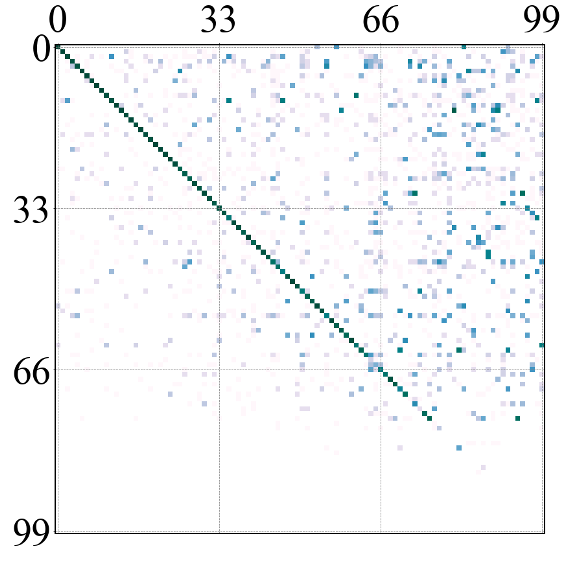}
        \label{fig:vis_cf_CB}
    }
    \subfloat[BS\cite{BS}]{
        \includegraphics[width=0.15\linewidth]
        {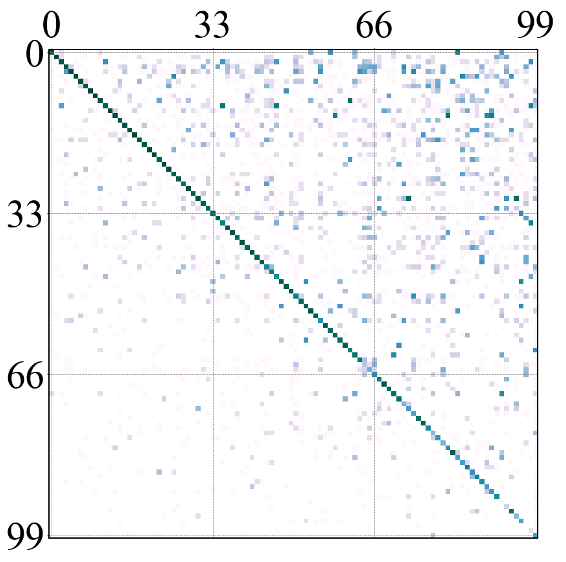}
        \label{fig:vis_cf_BC}
    }
    \subfloat[GCL\cite{GCL}]{
        \includegraphics[width=0.15\linewidth]
        {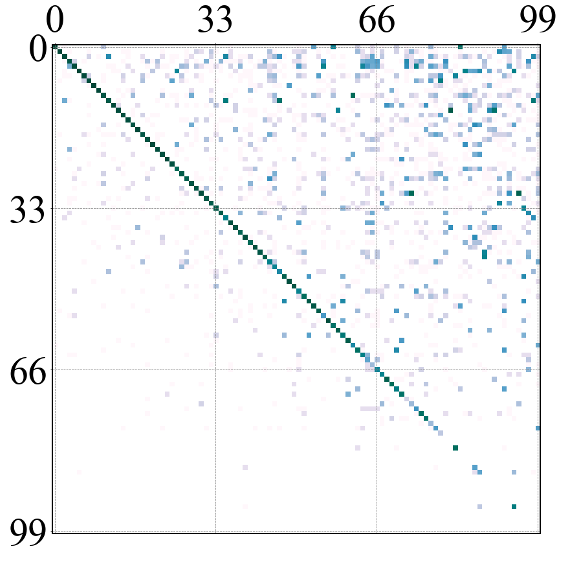}
        \label{fig:vis_cf_GCL}
    }
    \subfloat[NCL\cite{NCL}]{
        \includegraphics[width=0.15\linewidth]
        {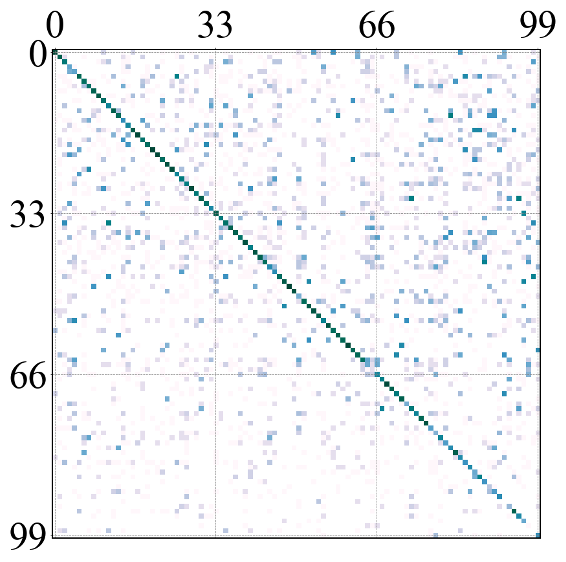}
        \label{fig:vis_cf_NCL}
    }
    \subfloat[{\name}]{
        \includegraphics[width=0.15\linewidth]
        {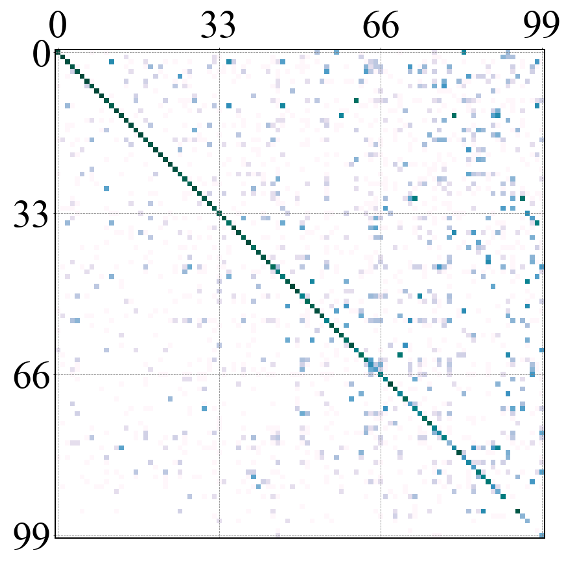}
        \label{fig:vis_cf_ECL}
    }
	\caption{Visualized $\log$-confusion matrix on CIFAR100-LT ($\gamma=100$). $x$-axis: ground truth. $y$-axis: predicted label. The deeper color indicates larger values. {\name} shows the best class accuracy and the most balanced misclassification distribution.}
	\label{fig:vis_cf}
	\vspace{-15pt}
\end{figure*}

\myparagraph{Do long-tail problems get alleviated?}
One of the primary goals of LTR is to improve performance in few-shot categories. Hence, we plot the confusion matrices on CIFAR100-LT. For better visualizations, we adopt logarithmic operations for all matrix values. In Fig.~\ref{fig:vis_cf}, the baseline (\ref{fig:vis_cf_CE}) prefers to train a trivial predictor, which simplifies images as many-shot labels to minimize the error rate. Several recent methods (\ref{fig:vis_cf_CB}-\ref{fig:vis_cf_NCL}) alleviate such issues to some extent. Compared to them, our proposal (\ref{fig:vis_cf_ECL}) shows the best accuracy (diagonal) and a more balanced misclassification distribution (non-diagonal). It firmly demonstrates our superiority in erasing the bias in LTR and our success in regularizing the few-shot classes.

\section{Conclusion}
\label{sec:conclusion}
This paper systematically analyzes the multi-expert framework in the long tail visual recognition, which trains several experts collaboratively to overcome the model preference for the majority and the high uncertainty on the minority. 
We point out that there is imbalanced knowledge transfer among experts' distillation, which leads to the inconspicuous improvement of collaborative learning on tail performance. A balanced distillation loss is proposed to improve the efficiency of collaborative learning by comparing two classifiers' predictions, which are supervised by different signals.
Furthermore, we claim that distillation at the feature level will greatly improve the feature quality and model performance. To learn representations more thoroughly, we integrate a contrastive proxy task and finally propose an effective collaborative learning framework, which helps the model extract robust features and learn meticulous distinguishing ability. We conduct both quantitative and qualitative experiments on four standard datasets to verify the superiority and effectiveness of ECL. Extensive experiments and visualizations demonstrate that ECL achieves state-of-the-art performance with better feature representations.

\bibliographystyle{IEEEtran}
\bibliography{ref}

\begin{thebibliography}{10}
\providecommand{\url}[1]{#1}
\csname url@samestyle\endcsname
\providecommand{\newblock}{\relax}
\providecommand{\bibinfo}[2]{#2}
\providecommand{\BIBentrySTDinterwordspacing}{\spaceskip=0pt\relax}
\providecommand{\BIBentryALTinterwordstretchfactor}{4}
\providecommand{\BIBentryALTinterwordspacing}{\spaceskip=\fontdimen2\font plus
\BIBentryALTinterwordstretchfactor\fontdimen3\font minus
  \fontdimen4\font\relax}
\providecommand{\BIBforeignlanguage}[2]{{%
\expandafter\ifx\csname l@#1\endcsname\relax
\typeout{** WARNING: IEEEtran.bst: No hyphenation pattern has been}%
\typeout{** loaded for the language `#1'. Using the pattern for}%
\typeout{** the default language instead.}%
\else
\language=\csname l@#1\endcsname
\fi
#2}}
\providecommand{\BIBdecl}{\relax}
\BIBdecl

\bibitem{ForestDet}
J.~Wu, L.~Song, Q.~Zhang, M.~Yang, and J.~Yuan, ``Forestdet: Large-vocabulary
  long-tailed object detection and instance segmentation,'' \emph{IEEE
  Transactions on Multimedia}, vol.~24, pp. 3693--3705, 2021.

\bibitem{VideoLT}
X.~Zhang, C.~Zhu, H.~Wu, Z.~Liu, and Y.~Xu, ``An imbalance compensation
  framework for background subtraction,'' \emph{IEEE Transactions on
  Multimedia}, vol.~19, no.~11, pp. 2425--2438, 2017.

\bibitem{LTReID}
P.~Wang, Z.~Zhao, F.~Su, and H.~Meng, ``Ltreid: Factorizable feature generation
  with independent components for long-tailed person re-identification,''
  \emph{IEEE Transactions on Multimedia}, 2022.

\bibitem{IPDLT}
M.~Ding, S.~Zhang, and J.~Yang, ``Improving pedestrian detection from a
  long-tailed domain perspective,'' in \emph{Proceedings of the 29th ACM
  International Conference on Multimedia}, 2021, pp. 2918--2926.

\bibitem{Imagenet}
O.~Russakovsky, J.~Deng, H.~Su, J.~Krause, S.~Satheesh, S.~Ma, Z.~Huang,
  A.~Karpathy, A.~Khosla, M.~Bernstein, A.~C. Berg, and L.~Fei-Fei, ``{ImageNet
  Large Scale Visual Recognition Challenge},'' \emph{IJCV}, vol. 115, no.~3,
  pp. 211--252, 2015.

\bibitem{COCO}
T.-Y. Lin, M.~Maire, S.~Belongie, J.~Hays, P.~Perona, D.~Ramanan,
  P.~Doll{\'a}r, and C.~L. Zitnick, ``Microsoft coco: Common objects in
  context,'' in \emph{ECCV}.\hskip 1em plus 0.5em minus 0.4em\relax Springer,
  2014, pp. 740--755.

\bibitem{PlaceLT}
B.~Zhou, A.~Lapedriza, A.~Khosla, A.~Oliva, and A.~Torralba, ``Places: A 10
  million image database for scene recognition,'' \emph{IEEE TPAMI}, 2017.

\bibitem{BBN}
B.~Zhou, Q.~Cui, X.-S. Wei, and Z.-M. Chen, ``Bbn: Bilateral-branch network
  with cumulative learning for long-tailed visual recognition,'' in
  \emph{CVPR}, 2020, pp. 9719--9728.

\bibitem{Bagoftricks}
Y.~Zhang, X.-S. Wei, B.~Zhou, and J.~Wu, ``Bag of tricks for long-tailed visual
  recognition with deep convolutional neural networks,'' in \emph{AAAI}, 2021,
  pp. 3447--3455.

\bibitem{Dynamicmix}
J.~Gao, J.~Chen, H.~Fu, and Y.-G. Jiang, ``Dynamic mixup for multi-label
  long-tailed food ingredient recognition,'' \emph{IEEE Transactions on
  Multimedia}, 2022.

\bibitem{PriorLT}
Z.~Xu, Z.~Chai, C.~Yuan \emph{et~al.}, ``Towards calibrated model for
  long-tailed visual recognition from prior perspective,'' \emph{NeurIPS},
  vol.~34, pp. 7139--7152, 2021.

\bibitem{FSA}
P.~Chu, X.~Bian, S.~Liu, and H.~Ling, ``Feature space augmentation for
  long-tailed data,'' in \emph{ECCV}.\hskip 1em plus 0.5em minus 0.4em\relax
  Springer, 2020, pp. 694--710.

\bibitem{M2m}
J.~Kim, J.~Jeong, J.~Shin \emph{et~al.}, ``M2m: Imbalanced classification via
  major-to-minor translation,'' in \emph{CVPR}, 2020, pp. 13\,896--13\,905.

\bibitem{CMO}
S.~Park, Y.~Hong, B.~Heo, S.~Yun, and J.~Y. Choi, ``The majority can help the
  minority: Context-rich minority oversampling for long-tailed
  classification,'' in \emph{CVPR}, 2022, pp. 6887--6896.

\bibitem{BS}
J.~Ren, C.~Yu, X.~Ma, H.~Zhao, S.~Yi \emph{et~al.}, ``Balanced meta-softmax for
  long-tailed visual recognition,'' \emph{NeurIPS}, vol.~33, pp. 4175--4186,
  2020.

\bibitem{LA}
A.~K. Menon, S.~Jayasumana, A.~S. Rawat, H.~Jain, A.~Veit, and S.~Kumar,
  ``Long-tail learning via logit adjustment,'' in \emph{ICLR}, 2021.

\bibitem{LADE}
Y.~Hong, S.~Han, K.~Choi, S.~Seo, B.~Kim, and B.~Chang, ``Disentangling label
  distribution for long-tailed visual recognition,'' in \emph{CVPR}.\hskip 1em
  plus 0.5em minus 0.4em\relax Computer Vision Foundation / {IEEE}, 2021, pp.
  6626--6636.

\bibitem{LFME}
L.~Xiang, G.~Ding, J.~Han \emph{et~al.}, ``Learning from multiple experts:
  Self-paced knowledge distillation for long-tailed classification,'' in
  \emph{ECCV}.\hskip 1em plus 0.5em minus 0.4em\relax Springer, 2020, pp.
  247--263.

\bibitem{RIDE}
X.~Wang, L.~Lian, Z.~Miao, Z.~Liu, and S.~Yu, ``Long-tailed recognition by
  routing diverse distribution-aware experts,'' in \emph{ICLR}, 2021.

\bibitem{NCL}
J.~Li, Z.~Tan, J.~Wan, Z.~Lei, and G.~Guo, ``Nested collaborative learning for
  long-tailed visual recognition,'' in \emph{CVPR}, 2022, pp. 6949--6958.

\bibitem{RTG}
Y.~Niu, L.~Chen, C.~Zhou, and H.~Zhang, ``Respecting transfer gap in knowledge
  distillation,'' in \emph{NeurIPS}, 2022.

\bibitem{MoCo}
K.~He, H.~Fan, Y.~Wu, S.~Xie, and R.~Girshick, ``Momentum contrast for
  unsupervised visual representation learning,'' in \emph{CVPR}, 2020, pp.
  9729--9738.

\bibitem{oversample1}
H.~Han, W.~Wang, B.~Mao \emph{et~al.}, ``Borderline-smote: {A} new
  over-sampling method in imbalanced data sets learning,'' in \emph{ICIC}, ser.
  Lecture Notes in Computer Science, vol. 3644.\hskip 1em plus 0.5em minus
  0.4em\relax Springer, 2005, pp. 878--887.

\bibitem{NCM}
B.~Kang, S.~Xie, M.~Rohrbach, Z.~Yan, A.~Gordo, J.~Feng, and Y.~Kalantidis,
  ``Decoupling representation and classifier for long-tailed recognition,'' in
  \emph{ICLR}, 2020.

\bibitem{RSG}
J.~Wang, T.~Lukasiewicz, X.~Hu, J.~Cai, and Z.~Xu, ``{RSG:} {A} simple but
  effective module for learning imbalanced datasets,'' in \emph{CVPR}.\hskip
  1em plus 0.5em minus 0.4em\relax Computer Vision Foundation / {IEEE}, 2021,
  pp. 3784--3793.

\bibitem{GeneticGAN}
J.~Hao, C.~Wang, G.~Yang, Z.~Gao, J.~Zhang, and H.~Zhang, ``Annealing genetic
  gan for imbalanced web data learning,'' \emph{IEEE Transactions on
  Multimedia}, vol.~24, pp. 1164--1174, 2021.

\bibitem{LDAM}
K.~Cao, C.~Wei, A.~Gaidon, N.~Arechiga, and T.~Ma, ``Learning imbalanced
  datasets with label-distribution-aware margin loss,'' \emph{NeurIPS},
  vol.~32, 2019.

\bibitem{MiSLAS}
Z.~Zhong, J.~Cui, S.~Liu, and J.~Jia, ``Improving calibration for long-tailed
  recognition,'' in \emph{CVPR}.\hskip 1em plus 0.5em minus 0.4em\relax
  Computer Vision Foundation / {IEEE}, 2021, pp. 16\,489--16\,498.

\bibitem{SimCLR}
T.~Chen, S.~Kornblith, M.~Norouzi, and G.~Hinton, ``A simple framework for
  contrastive learning of visual representations,'' in \emph{ICML}.\hskip 1em
  plus 0.5em minus 0.4em\relax PMLR, 2020, pp. 1597--1607.

\bibitem{SSLMM}
Z.~Tao, X.~Liu, Y.~Xia, X.~Wang, L.~Yang, X.~Huang, and T.-S. Chua,
  ``Self-supervised learning for multimedia recommendation,'' \emph{IEEE
  Transactions on Multimedia}, 2022.

\bibitem{SSP}
Y.~Yang, Z.~Xu \emph{et~al.}, ``Rethinking the value of labels for improving
  class-imbalanced learning,'' \emph{NeurIPS}, vol.~33, pp. 19\,290--19\,301,
  2020.

\bibitem{HybirdSC}
P.~Wang, K.~Han, X.-S. Wei, L.~Zhang, and L.~Wang, ``Contrastive learning based
  hybrid networks for long-tailed image classification,'' in \emph{CVPR}, 2021,
  pp. 943--952.

\bibitem{PaCo}
J.~Cui, Z.~Zhong, S.~Liu, B.~Yu, and J.~Jia, ``Parametric contrastive
  learning,'' in \emph{ICCV}, 2021, pp. 715--724.

\bibitem{TSC}
T.~Li, P.~Cao, Y.~Yuan, L.~Fan, Y.~Yang, R.~S. Feris, P.~Indyk, and D.~Katabi,
  ``Targeted supervised contrastive learning for long-tailed recognition,'' in
  \emph{CVPR}, 2022, pp. 6918--6928.

\bibitem{BCL}
J.~Zhu, Z.~Wang, J.~Chen, Y.-P.~P. Chen, and Y.-G. Jiang, ``Balanced
  contrastive learning for long-tailed visual recognition,'' in \emph{CVPR},
  2022, pp. 6908--6917.

\bibitem{DAP}
M.~A. Jamal, M.~Brown, M.-H. Yang, L.~Wang, and B.~Gong, ``Rethinking
  class-balanced methods for long-tailed visual recognition from a domain
  adaptation perspective,'' in \emph{CVPR}, 2020, pp. 7610--7619.

\bibitem{ISFDA}
X.~Li, J.~Li, L.~Zhu, G.~Wang, and Z.~Huang, ``Imbalanced source-free domain
  adaptation,'' in \emph{Proceedings of the 29th ACM International Conference
  on Multimedia}, 2021, pp. 3330--3339.

\bibitem{GCL}
M.~Li, Y.-m. Cheung, Y.~Lu \emph{et~al.}, ``Long-tailed visual recognition via
  gaussian clouded logit adjustment,'' in \emph{CVPR}, 2022, pp. 6929--6938.

\bibitem{CB}
Y.~Cui, M.~Jia, T.-Y. Lin, Y.~Song, and S.~Belongie, ``Class-balanced loss
  based on effective number of samples,'' in \emph{CVPR}, 2019, pp. 9268--9277.

\bibitem{LTDA}
Z.~Peng, W.~Huang, Z.~Guo, X.~Zhang, J.~Jiao, and Q.~Ye, ``Long-tailed
  distribution adaptation,'' in \emph{Proceedings of the 29th ACM International
  Conference on Multimedia}, 2021, pp. 3275--3282.

\bibitem{EqLoss}
J.~Sun, W.~Yang, J.-H. Xue, and Q.~Liao, ``An equalized margin loss for face
  recognition,'' \emph{IEEE Transactions on Multimedia}, vol.~22, no.~11, pp.
  2833--2843, 2020.

\bibitem{LTR-WD}
S.~Alshammari, Y.-X. Wang, D.~Ramanan, and S.~Kong, ``Long-tailed recognition
  via weight balancing,'' in \emph{CVPR}, 2022, pp. 6897--6907.

\bibitem{TADE}
Y.~Zhang, B.~Hooi, L.~Hong, and J.~Feng, ``Self-supervised aggregation of
  diverse experts for test-agnostic long-tailed recognition,'' \emph{NeurIPS},
  vol.~35, pp. 34\,077--34\,090, 2022.

\bibitem{CBD}
A.~Iscen, A.~Araujo, B.~Gong, and C.~Schmid, ``Class-balanced distillation for
  long-tailed visual recognition,'' in \emph{BMVC}.\hskip 1em plus 0.5em minus
  0.4em\relax {BMVA} Press, 2021, p. 165.

\bibitem{ACE}
J.~Cai, Y.~Wang, J.-N. Hwang \emph{et~al.}, ``Ace: Ally complementary experts
  for solving long-tailed recognition in one-shot,'' in \emph{ICCV}, 2021, pp.
  112--121.

\bibitem{KD}
G.~Hinton, O.~Vinyals, J.~Dean \emph{et~al.}, ``Distilling the knowledge in a
  neural network,'' \emph{arXiv preprint arXiv:1503.02531}, vol.~2, no.~7,
  2015.

\bibitem{DiVE}
Y.-Y. He, J.~Wu, X.-S. Wei \emph{et~al.}, ``Distilling virtual examples for
  long-tailed recognition,'' in \emph{ICCV}, 2021, pp. 235--244.

\bibitem{SSD}
T.~Li, L.~Wang, and G.~Wu, ``Self supervision to distillation for long-tailed
  visual recognition,'' in \emph{ICCV}, 2021, pp. 630--639.

\bibitem{Landscape}
H.~Li, Z.~Xu, G.~Taylor, C.~Studer, and T.~Goldstein, ``Visualizing the loss
  landscape of neural nets,'' in \emph{NeurIPS}, 2018.

\bibitem{Focal}
T.-Y. Lin, P.~Goyal, R.~Girshick, K.~He, and P.~Doll{\'a}r, ``Focal loss for
  dense object detection,'' in \emph{ICCV}, 2017, pp. 2980--2988.

\bibitem{CausalNorm}
K.~Tang, J.~Huang, and H.~Zhang, ``Long-tailed classification by keeping the
  good and removing the bad momentum causal effect,'' \emph{NeurIPS}, vol.~33,
  pp. 1513--1524, 2020.

\bibitem{DRO}
D.~Samuel, G.~Chechik \emph{et~al.}, ``Distributional robustness loss for
  long-tail learning,'' in \emph{ICCV}, 2021, pp. 9495--9504.

\bibitem{IDR}
S.~Yu, J.~Guo, R.~Zhang, Y.~Fan, Z.~Wang, and X.~Cheng, ``A re-balancing
  strategy for class-imbalanced classification based on instance difficulty,''
  in \emph{CVPR}, 2022, pp. 70--79.

\bibitem{Cifar}
A.~Krizhevsky, G.~Hinton \emph{et~al.}, ``Learning multiple layers of features
  from tiny images,'' 2009.

\bibitem{iNaturalist}
G.~Van~Horn, O.~Mac~Aodha, Y.~Song, Y.~Cui, C.~Sun, A.~Shepard, H.~Adam,
  P.~Perona, and S.~Belongie, ``The inaturalist species classification and
  detection dataset,'' in \emph{CVPR}, 2018, pp. 8769--8778.

\bibitem{OLTR}
Z.~Liu, Z.~Miao, X.~Zhan, J.~Wang, B.~Gong, and S.~X. Yu, ``Large-scale
  long-tailed recognition in an open world,'' in \emph{CVPR}, 2019.

\bibitem{DisAlign}
S.~Zhang, Z.~Li, S.~Yan, X.~He, and J.~Sun, ``Distribution alignment: A unified
  framework for long-tail visual recognition.'' in \emph{CVPR}, 2021.

\bibitem{CKT}
S.~Parisot, P.~M. Esperan{\c{c}}a, S.~McDonagh, T.~J. Madarasz, Y.~Yang, and
  Z.~Li, ``Long-tail recognition via compositional knowledge transfer,'' in
  \emph{CVPR}, 2022, pp. 6939--6948.

\bibitem{ECE}
A.~Ashukha, A.~Lyzhov, D.~Molchanov, and D.~Vetrov, ``Pitfalls of in-domain
  uncertainty estimation and ensembling in deep learning,'' in \emph{ICLR},
  2020.

\bibitem{CosLearning}
I.~Loshchilov and F.~Hutter, ``{SGDR}: Stochastic gradient descent with warm
  restarts,'' in \emph{ICLR}, 2017.

\bibitem{Cutout}
T.~DeVries, G.~W. Taylor \emph{et~al.}, ``Improved regularization of
  convolutional neural networks with cutout,'' \emph{arXiv preprint
  arXiv:1708.04552}, 2017.

\bibitem{Autoaug}
Y.~Chen, Y.~Li, T.~Kong, L.~Qi, R.~Chu, L.~Li, and J.~Jia, ``Scale-aware
  automatic augmentation for object detection,'' in \emph{CVPR}, 2021.

\bibitem{ResNet}
K.~He, X.~Zhang, S.~Ren, and J.~Sun, ``Deep residual learning for image
  recognition,'' in \emph{CVPR}, 2016, pp. 770--778.

\bibitem{Randaugment}
E.~D. Cubuk, B.~Zoph, J.~Shlens, and Q.~V. Le, ``Randaugment: Practical
  automated data augmentation with a reduced search space,'' in \emph{CVPR
  workshops}, 2020, pp. 702--703.

\bibitem{ResNeXt}
S.~Xie, R.~Girshick, P.~Doll{\'a}r, Z.~Tu, and K.~He, ``Aggregated residual
  transformations for deep neural networks,'' in \emph{CVPR}, 2017, pp.
  1492--1500.

\bibitem{tsne}
L.~Van~der Maaten and G.~Hinton, ``Visualizing data using t-sne,''
  \emph{Journal of machine learning research}, vol.~9, no.~11, 2008.

\end{thebibliography}
\clearpage

\end{document}